\documentclass[runningheads]{llncs}

\usepackage{eccv}

\usepackage{wrapfig}
\usepackage{multirow}
\usepackage{pifont}
\usepackage[table]{xcolor}
\usepackage{enumitem}

\usepackage{eccvabbrv}

\usepackage{graphicx}
\usepackage{booktabs}

\usepackage[accsupp]{axessibility}  % Improves PDF readability for those with disabilities.

\usepackage{hyperref}

\usepackage{orcidlink}

\begin{document}

\title{Symbiotic-MoE: Unlocking the Synergy between Generation and Understanding}

\titlerunning{Symbiotic-MoE}

\author{
    Xiangyue Liu\inst{1}, 
    Zijian Zhang\inst{2}, 
    Miles Yang\inst{2}, 
    Zhao Zhong\inst{2}, \\[1mm] 
    Liefeng Bo\inst{2}, 
    Ping Tan\inst{1}\thanks{Corresponding author.} 
}
\authorrunning{X. Liu et al.}

\institute{
    \vspace{-2mm} 
    \textsuperscript{1}HKUST \quad \textsuperscript{2}Tencent Hunyuan \\
    % \vspace{2mm} 
    % \vspace{-1mm}
}

% % TODO FINAL: Replace with an abbreviated list of authors.

% % TODO FINAL: Replace with your institution list.
% Springer Heidelberg, Tiergartenstr.~17, 69121 Heidelberg, Germany
% % \url{http://www.springer.com/gp/computer-science/lncs} \and
% % \email{\{abc,lncs\}@uni-heidelberg.de}}

\maketitle

\begin{figure}[h!] 
    \centering
    \vspace{-2em} 
    \includegraphics[width=0.90\textwidth]{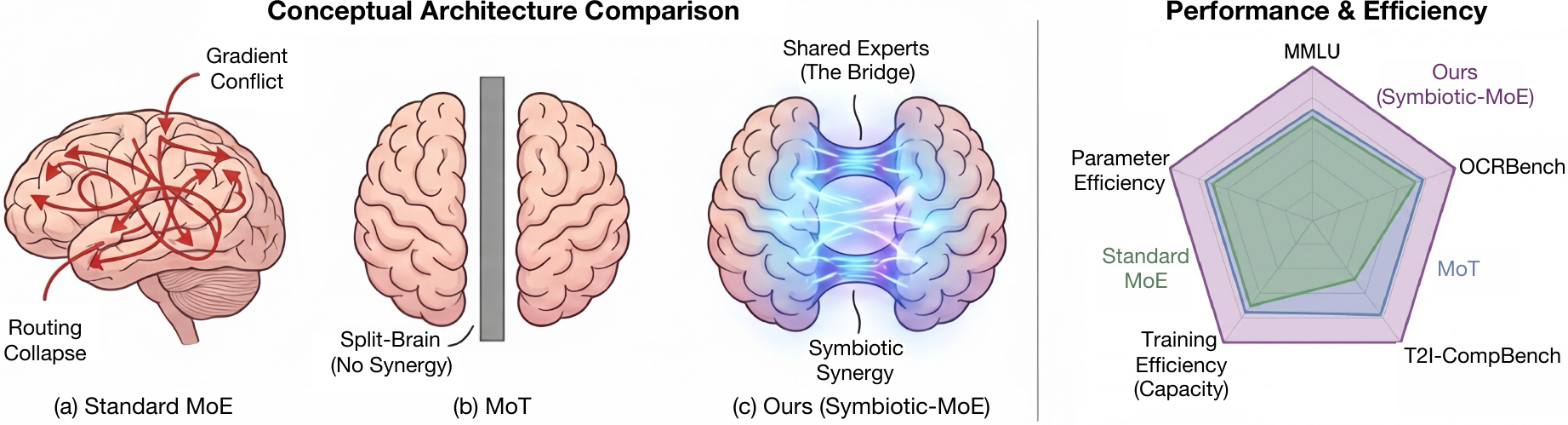}
    \caption{\textbf{Teaser.} 
    We visualize the evolution of training paradigms: moving from (a) destructive interference and (b) the ``Split-Brain'' compromise to (c) symbiotic synergy. As shown in the radar chart, our framework achieves holistic superiority, simultaneously boosting generation and understanding capabilities with zero-parameter overhead and maximal parameter efficiency.}
    \label{fig:teaser}
    \vspace{-3em} 
\end{figure}

\begin{abstract}
    Empowering Large Multimodal Models (LMMs) with image generation often leads to catastrophic forgetting in understanding tasks due to severe gradient conflicts. While existing paradigms like Mixture-of-Transformers (MoT) mitigate this conflict through structural isolation, they fundamentally sever cross-modal synergy and suffer from capacity fragmentation. In this work, we present Symbiotic-MoE, a unified pre-training framework that resolves task interference within a native multimodal Mixture-of-Experts (MoE) Transformers architecture with zero-parameter overhead. We first identify that standard MoE tuning leads to routing collapse, where generative gradients dominate expert utilization. To address this, we introduce Modality-Aware Expert Disentanglement, which partitions experts into task-specific groups while utilizing shared experts as a multimodal semantic bridge. Crucially, this design allows shared experts to absorb fine-grained visual semantics from generative tasks to enrich textual representations. To optimize this, we propose a Progressive Training Strategy featuring differential learning rates and early-stage gradient shielding. This mechanism not only shields pre-trained knowledge from early volatility but eventually transforms generative signals into constructive feedback for understanding. Extensive experiments demonstrate that Symbiotic-MoE achieves rapid generative convergence while unlocking cross-modal synergy, boosting inherent understanding with remarkable gains on MMLU and OCRBench.
\end{abstract}

\section{Introduction}
\label{sec:intro}

\begin{figure}[t] 
    \centering
    \includegraphics[width=1.0\textwidth]{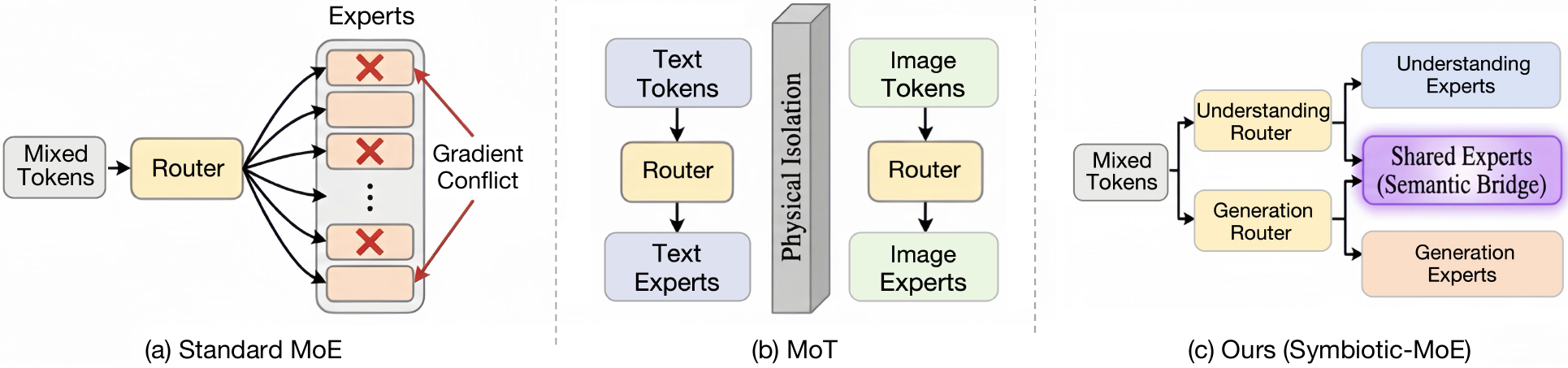}
    \caption{\textbf{Comparison of Architectures.} (a) Standard MoE suffers from routing collapse due to multi-task gradient conflicts (like cognitive dissonance). (b) MoT avoids conflict via physical isolation but induces a ``Split-Brain'' dilemma, which structurally hinders cross-modal synergy and knowledge transfer. (c) Our Symbiotic-MoE introduces shared experts as a semantic bridge, enabling co-evolution of generation and understanding within a unified, parameter-efficient architecture.}
    \label{fig:fig2}
      \vspace{-9pt}
    % \vspace{-2em} 
\end{figure}

The convergence of perception and creation within a single cognitive system has long been a holy grail in the pursuit of Artificial General Intelligence (AGI). In recent years, Large Multimodal Models (LMMs) have made remarkable strides in understanding the visual world~\cite{achiam2023gpt, team2023gemini, liu2023visual}. However, enabling these models to also generate visual content that transforms them into true ``Any-to-Any'' omni-modal foundation models remains a formidable challenge. Recent advances have attempted to unify vision and language through diverse paradigms, ranging from discrete tokenization~\cite{team2024chameleon, sun2023emu} to continuous feature alignment~\cite{zhou2024transfusion, chen2025janus, xiao2025omnigen}. Despite these architectural innovations, a critical optimization bottleneck persists: extending a pre-trained understanding model with generative capabilities often triggers severe catastrophic forgetting.

This phenomenon stems from the intrinsic stability-plasticity dilemma~\cite{mccloskey1989catastrophic, parisi2019continual}. Visual understanding tasks, which require the model to converge onto precise semantic representations (many-to-one mapping), fundamentally conflict with the high-entropy, divergent nature of generative tasks (one-to-many mapping). When naively co-trained, the high-variance gradients from the generative objective tend to overwhelm the established optimization landscape of the understanding task, washing away pre-trained knowledge. Consequently, current unified models often face a zero-sum game: significant improvements in generation quality typically come at the expense of understanding retention, or necessitate complex, multi-stage training strategies to mitigate interference.

To mitigate this interference, prior arts have predominantly resorted to structural isolation. Approaches like Mixture-of-Transformers (MoT)\cite{liang2025mixtureoftransformers, deng2025bagel} or adapter-based methods\cite{wu2025janus, xie2024show, dong2023dreamllm} physically decouple the parameters for different modalities. While effective at preserving stability, this ``divide-and-conquer'' strategy comes at a steep cost: it often introduces significant parameter overhead, increases inference latency, and most critically, severs the semantic connectivity between modalities. By segregating the processing pathways, these methods forfeit the potential synergy where generative feedback could refine understanding—a cognitive mechanism inherent in biological intelligence.

In this work, we ask a fundamental question: \textit{Can we achieve the stability of isolated architectures while retaining the synergistic benefits of a unified model, without any parameter overhead?} To answer this, we present Symbiotic-MoE, a novel pre-training framework that harmonizes generation and understanding within a native sparse Mixture-of-Experts (MoE) Transformers architecture.

Our key insight is that task interference is not a failure of the unified architecture itself, but a failure of routing dynamics. We observe that standard MoE training leads to routing collapse, where generative tokens monopolize the experts, starving the understanding task. To resolve this, we propose Modality-Aware Expert Disentanglement. Instead of physical separation, we logically partition the expert space into task-specific groups based on their pre-trained activation patterns. Crucially, we introduce shared experts to act as a multimodal semantic bridge. This private-shared design allows task-specific experts to specialize in conflicting objectives, while shared experts facilitate deep cross-modal alignment, turning the conflict into symbiosis.

Furthermore, we recognize that architecture alone is insufficient to tame the volatile training dynamics of generative adaptation. We thus introduce the Progressive Training Strategy. We devise a Knowledge-Inherited Initialization to eliminate the cold-start penalty and a Differential Learning Rate schedule to balance the update pace of different modules. Most notably, we propose a Warmup Gradient Shielding mechanism. This strategy temporarily blocks noisy generative gradients from updating shared experts during the early unstable phase, protecting the foundational knowledge until the generative module matures.

Our contributions are summarized as follows:
\begin{itemize}
    \item We propose Symbiotic-MoE, a zero-overhead framework that resolves routing collapse in unified multimodal pre-training. It introduces Modality-Aware Expert Disentanglement to resolve resource contention and incorporates Shared Experts as a multimodal bridge to enable seamless cross-modal alignment.
    \item We introduce a Progressive Training Strategy that harmonizes the conflicting optimization dynamics. By employing Differential Learning Rates and Warmup Gradient Shielding, we effectively navigate the stability-plasticity dilemma, protecting pre-trained knowledge while unlocking generative capabilities.
    \item Extensive experiments demonstrate that our method not only achieves rapid generative convergence but also boosts understanding capabilities. This provides empirical validation for the symbiotic hypothesis: generative training, when properly orchestrated, can reciprocally enhance multimodal understanding.
\end{itemize}
\section{Related Work}
\label{sec:related}

\subsection{Unified Multimodal Understanding and Generation}
The evolution of Large Multimodal Models (LMMs) has progressed from perception centric systems~\cite{lu2024deepseek, wang2024qwen2, chen2024internvl, team2023gemini} to unified ``Any-to-Any'' frameworks~\cite{tang2023any, wu2024next}. Early attempts cascaded LLMs with diffusion models~\cite{wu2023visual, koh2023generating}, limiting end-to-end synergy. To achieve native unification, pioneers like Chameleon~\cite{team2024chameleon} and Emu3~\cite{wang2024emu3} adopted discrete tokenization, formulating image generation as auto-regressive next-token prediction. Conversely, continuous approaches like Transfusion~\cite{zhou2024transfusion} and Show-o~\cite{xie2024show} integrated diffusion or flow-matching objectives directly into the transformers backbone to improve fidelity. More recently, to mitigate modality interference, Janus~\cite{wu2025janus} explored decoupled visual encodings, while OmniGen~\cite{xiao2025omnigen} pushed for a unified diffusion transformer. Despite these strides, training a single backbone for both tasks remains unstable. The divergent nature of generation (one-to-many) and the convergent nature of understanding (many-to-one) create a fundamental optimization paradox~\cite{mccloskey1989catastrophic, parisi2019continual}. This conflict often leads to a ``seesaw effect'', where improving generative plasticity inevitably degrades understanding stability.

\subsection{Multimodal Mixture-of-Experts (MoE)} 
Sparse Mixture-of-Experts (MoE)~\cite{shazeer2017outrageously, lepikhin2020gshard, fedus2022switch} Transformers has emerged as a promising solution to scale model capacity without inflating inference costs, widely adopted in LLMs~\cite{du2022glam, zoph2022st, gu2025delta, lv2025coupling}, most notably exemplified by milestones such as Mixtral~\cite{jiang2024mixtral}, DeepSeek-MoE~\cite{dai2024deepseekmoe} and Qwen3~\cite{yang2025qwen3}. 
In the multimodal domain, sparse architectures have proliferated to handle high-resolution inputs efficiently~\cite{team2025kimi, luo2025mono, wang2025moiie, xu2026tag}. Beyond foundational works like MoE-LLaVA~\cite{lin2026moe}, V-MoE~\cite{riquelme2021scaling}, and MM1~\cite{mckinzie2024mm1}, recent advances including DeepSeek-V3~\cite{liu2024deepseek}, MM1.5~\cite{zhang2024mm1} further demonstrate the scalability of experts in processing complex modalities. 
However, the application of MoE in unified generation and understanding remains underexplored. Standard routing mechanisms often fail in this context because the gradients from generative losses are significantly larger and noisier than those from understanding losses. This leads to a routing collapse phenomenon, where experts are overwhelmingly co-opted by the generative task, leaving the understanding capability starved of capacity. Our work is among the first to address this imbalance within a native MoE framework, turning the sparsity from a scaling tool into a mechanism for task disentanglement.

\subsection{Task Interference and Resolution Strategies}
Catastrophic forgetting~\cite{chen2024mofo, zhang2025metagdpo, zhang2025reinforcement, chen2025mol} arising from gradient conflicts remains a primary bottleneck in extending VLMs with generative capabilities. Prior works mitigate this via two dominant paradigms: Additive Adaptation and Structural Isolation. Additive approaches freeze the pre-trained backbone and append auxiliary modules, such as complex adapters or side-networks~\cite{hu2022lora, zhang2023adding, ye2023ip, gao2023llama, wu2024next}. While effective for stability, they intrinsically limit deep cross-modal interaction and incur non-negligible inference latency. Conversely, Structural Isolation methods, exemplified by MoT and recent sparse variants~\cite{liang2025mixtureoftransformers, deng2025bagel, wang2025hbridge}, physically decouple processing pathways for different modalities. Although this avoids interference, it enforces a ``Split-Brain'' architecture that severs potential synergy and suffers from capacity fragmentation. In contrast, Symbiotic-MoE proposes a Modality-Aware Expert Disentanglement strategy. We achieve the best of both worlds: the specialized optimization of decoupled architectures without parameter overhead, and the deep fusion benefits of a unified model via shared experts. By orchestrating gradient flow rather than physically separating parameters, we resolve interference while fostering positive cross-modal transfer.

\section{Method}
\label{sec:method}
In this section, we present \textbf{Symbiotic-MoE}, a unified pre-training framework designed to harmonize the conflicting objectives of visual generation and multimodal understanding within a single sparse architecture. Unlike prior approaches such as MoT~\cite{liang2025mixtureoftransformers, deng2025bagel} that enforce strict structural isolation, our method operates on the principle of \textit{co-evolution}. 

\begin{figure}[t] 
    \centering
    % \vspace{-1.5em} 
    \includegraphics[width=1.0\textwidth]{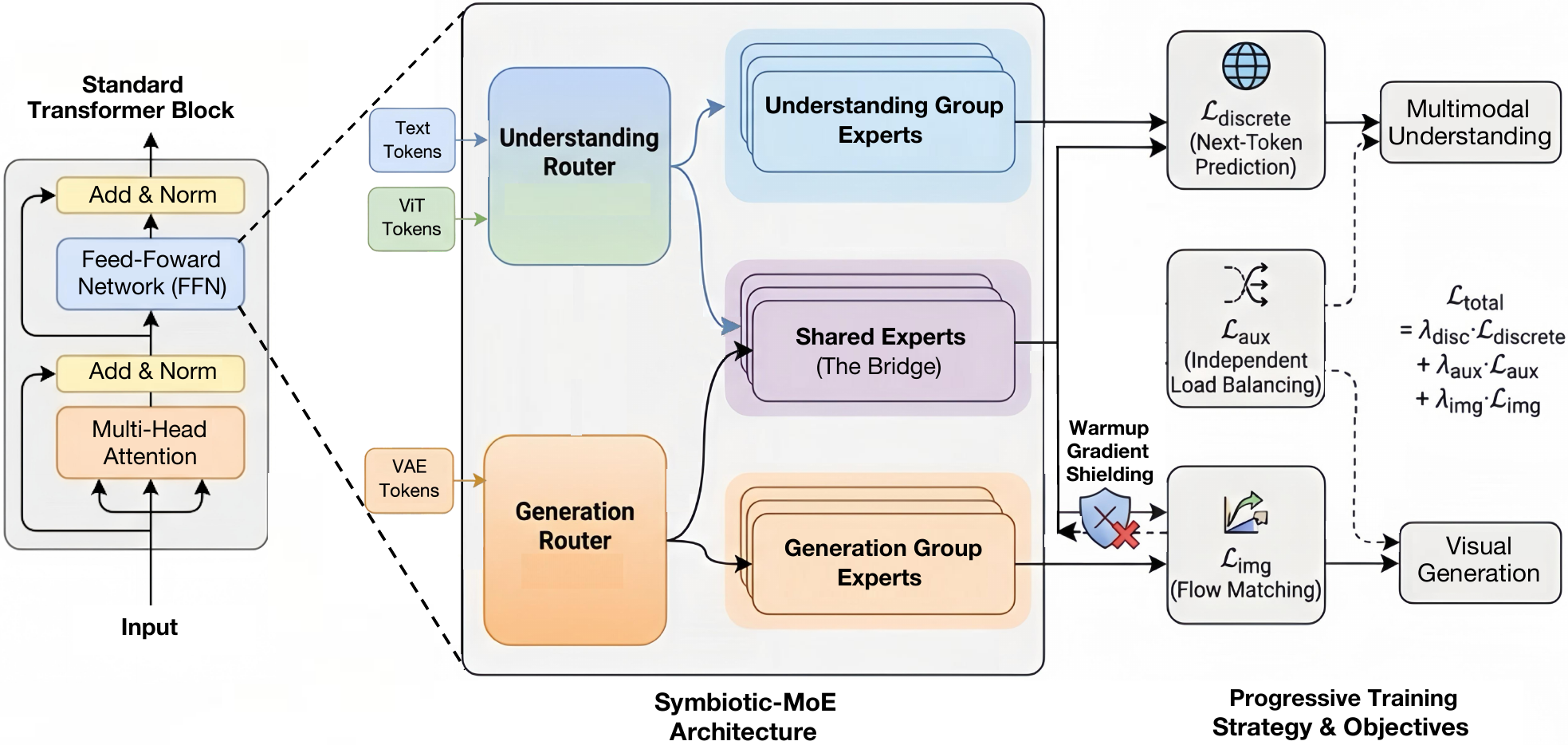}
    \caption{\textbf{Overview of Symbiotic-MoE.} We re-architect the Transformer FFN layer to resolve task interference. Our framework features: (1) \textit{Modality-Aware Expert Disentanglement}, where input tokens are routed to specialized Understanding (blue) or Generation (orange) experts via decoupled routers; (2) A \textit{Shared Expert Bridge} (purple), which processes all tokens to enforce semantic alignment and prevent modal isolation. The entire framework is optimized via \textit{Progressive Training Strategy}, harmonizing generation and understanding within a unified Transformer backbone.}
    \label{fig:pipeline}
    \vspace{-1em} 
\end{figure}

As illustrated in Fig.~\ref{fig:pipeline}, we address the catastrophic forgetting and routing collapse issues through two core components: 
(1) \textbf{Symbiotic-MoE Architecture} (Sec.~\ref{sec:architecture}), which employs \textit{Modality-Aware Expert Disentanglement} to partition experts into task-specific groups, bridged by \textit{Shared Experts} for cross-modal alignment, alongside \textit{Knowledge-Inherited Initialization} for a zero-cold-start transition; and 
(2) \textbf{Progressive Training Strategy} (Sec.~\ref{sec:progressive-training}), which resolves the stability-plasticity dilemma by orchestrating update dynamics through \textit{Differential Learning Rates} and \textit{Warmup Gradient Shielding}.

\subsection{Preliminaries}

\noindent\textbf{Sparse Mixture-of-Experts (MoE).} 
We adopt the standard sparse MoE formulation where the dense Feed-Forward Network (FFN) in a Transformer block is replaced by a set of experts $\mathcal{E} = \{E_i\}_{i=1}^N$. For an input $\mathbf{x}$, the router selects top-$k$ experts (denoted by indices $\mathcal{I}$) based on gating scores. The output is the weighted sum of the selected experts:
\begin{equation}
    \mathbf{y} = \sum_{i \in \mathcal{I}} g_i(\mathbf{x}) \cdot E_i(\mathbf{x})
\end{equation}
where $g_i(\mathbf{x})$ is the softmax-normalized routing weight for the chosen expert $i$. This architecture allows the model to scale capacity while maintaining constant inference costs.

\medskip\noindent\textbf{Conflicts in Co-training.} 
Naively integrating generative objectives into a pre-trained MoE-based VLM precipitates a severe stability-plasticity dilemma~\cite{mccloskey1989catastrophic, parisi2019continual}. The aggressive, high-magnitude gradients required for learning pixel-level synthesis inevitably overwhelm the converged optimization landscape of the understanding task. This gradient interference causes experts to drift uncontrollably toward the generative manifold, erasing pre-trained semantic structures and triggering catastrophic forgetting. Detailed visualization of the conflicts in 
Supplementary Material Sec. \hyperref[sec:supp_conflicts]{A}.
% Supplementary Material Sec. \textcolor{red}{A}.

\subsection{Symbiotic-MoE Architecture}
\label{sec:architecture}

\noindent\textbf{Modality-Aware Expert Disentanglement.}
To resolve the conflict between retaining pre-trained knowledge and learning new generative capabilities, we propose a specialized grouping strategy derived from the intrinsic activation patterns of the experts.

\smallskip\noindent\textit{Expert Role Analysis.} 
Instead of random grouping, we conduct a data-driven analysis to identify the role of each expert in the pre-trained VLM. Specifically, we perform inference on two representative benchmarks: MMLU~\cite{hendrycks2020measuring} (for linguistic Text tokens) and OCRBench~\cite{liu2024ocrbench} (for discriminative ViT tokens). We calculate the activation frequency of all 128 experts across all 47 MoE layers. By ranking experts based on their cumulative activation rates for Text and ViT tokens, we identify the core experts that are essential for the model's fundamental capabilities by modality. Detailed visualizations of activation landscapes are provided in Supplementary Material Sec. \hyperref[sec:supp_arch]{A}. 
% Supplementary Material Sec. \textcolor{red}{A}.

\smallskip\noindent\textit{Hard Routing Strategy.} 
Based on the frequency profiling, we implement a hard routing split. For each layer, we designate the top 96 experts with the highest relevance to original modalities (Text and ViT) as the \textit{understanding group}, ensuring that the vast majority of pre-trained knowledge is preserved (stability). The remaining 32 experts in each layer, which are less frequently activated, are repurposed as the \textit{generation group} assigned to the VAE features (plasticity). During training, tokens are routed to their respective groups based on their modality type. Accordingly, the parameters of the original router are sliced and assigned to initialize the specific routers for each group, ensuring a warm start.

\smallskip\noindent\textit{The Bridge: Shared Experts.} 
A critical distinction of our Symbiotic-MoE from prior hard-routing methods (\eg, MoT~\cite{liang2025mixtureoftransformers, deng2025bagel}) is the preservation of shared experts. We retain the shared experts in each layer to process \textit{all} tokens, regardless of whether they are Text, ViT, or VAE. This design allows the shared experts to function as a multimodal bridge, facilitating information exchange between the decoupled expert groups. It ensures that the generative process remains semantically aligned with the understanding representations, enabling cross-modal synergy rather than isolation.

\medskip\noindent\textbf{Knowledge-Inherited Initialization.}
\label{sec:initilization}
Transitioning from a unified MoE to a disentangled architecture typically incurs a cold-start penalty if the new components are randomly initialized. To mitigate this and ensure a seamless adaptation, we propose a \textit{Knowledge-Inherited Initialization} strategy that transfers the learned routing priors and expert capabilities from the pre-trained VLM to our Symbiotic-MoE.

\smallskip\noindent\textit{Expert and Shared Weight Inheritance.}
Since our architecture retains the physical structure of the experts, we directly initialize the parameters of the partitioned understanding and generation groups using the weights from their corresponding expert indices in the original VLM. Similarly, the shared experts, which serve as the multimodal bridge, inherit their parameters directly from the pre-trained shared experts. This ensures that the foundational knowledge stored within the VLM is preserved intact.

\smallskip\noindent\textit{Router Weight Slicing.}
The most critical challenge lies in initializing the two newly decoupled routers (one for understanding group, one for generation group) from the single original router. Random initialization would destroy the learned mapping between tokens and their preferred experts, leading to immediate performance collapse. To address this, we introduce \textit{Router Weight Slicing}. Let $\mathbf{W}_r \in \mathbb{R}^{d \times N}$ denote the projection matrix of the original router, where $N=128$ is the total number of experts. For a specific group $g \in \{\text{und}, \text{gen}\}$ containing a subset of expert indices $\mathcal{I}_g$, we construct the new router weight $\mathbf{W}_r^g$ by slicing the columns of $\mathbf{W}_r$ corresponding to $\mathcal{I}_g$. Mathematically, $\mathbf{W}_r^g = \mathbf{W}_r[:, \mathcal{I}_g]$. This operation effectively preserves the routing priors: a token that originally preferred Expert $i$ will still produce a high gating score for Expert $i$ in the new sub-router. This strategy achieves a zero-cold-start, allowing the model to maintain near-original understanding performance at iteration zero, as in Table~\ref{tab:ablation_grouping}.

\subsection{Progressive Training Strategy}
\label{sec:progressive-training}
Even with a disentangled architecture, the simultaneous optimization of a converged VLM and a newly added generative module presents a significant challenge due to their disparate optimization landscapes. To harmonize these conflicting dynamics, we propose a progressive training strategy.

\medskip\noindent\textbf{Differential Learning Rates.}
A unified learning rate is suboptimal for the Symbiotic-MoE due to the maturity gap between modules. The pre-trained VLM components (Text/ViT experts and routers) reside in a sharp local minimum where a high learning rate (\eg, $1e^{-4}$) triggers immediate divergence and catastrophic forgetting. Conversely, the newly initialized generative components (VAE experts and routers) require a larger step size to escape their initial random state and learn effective representations. To resolve this conflict, we implement a Differential Learning Rate schedule. We assign a larger learning rate ($1e^{-4}$) exclusively to the generation-centric components (VAE experts and routers) to accelerate their convergence. Simultaneously, we apply a conservative learning rate ($1e^{-6}$) to the understanding-centric components (Text/ViT experts, routers, and shared experts) to perform fine-grained adaptation without destroying their pre-trained priors. This differential optimization ensures that each module evolves at its appropriate pace. 

\medskip\noindent\textbf{Warmup Gradient Shielding.}
Initial phase of co-training is highly volatile. The generative module produces high variance gradients that can catastrophically distort the well-tuned semantic features within the shared experts. To prevent this, we introduce a temporal gradient shielding mechanism. Specifically, during the warmup period, while VAE tokens are permitted to forward-propagate through shared experts to leverage pre-trained features, we enforce a stop-gradient operation on the backward pass. This strategy acts as a protective buffer, ensuring that shared experts are updated solely by stable Text and ViT gradients while the generative module is in its chaotic early learning phase. Crucially, this shielding is transient. Once the generative optimization trajectory stabilizes after warmup iterations, we remove the constraint to enable full bidirectional gradient flow, thereby facilitating deep cross-modal fusion through shared experts. Concurrently, to counterbalance the update magnitude disparity arising from our differential learning rates ($1e^{-4}$ \vs $1e^{-6}$), we apply a gradient scaling factor of 0.1 to generative tokens within the shared experts, preventing aggressive generative signals from monopolizing the shared experts semantic space.

\medskip\noindent\textbf{Objective Functions.}
Our training objective is designed to strictly preserve the original VLM optimization landscape while integrating generative capabilities. The final loss function extends the original VLM objectives by incorporating the image generation term:
\begin{equation}
    \mathcal{L}_{total} = \lambda_{disc} \cdot \mathcal{L}_{discrete} + \lambda_{aux} \cdot \mathcal{L}_{aux} + \lambda_{img} \cdot \mathcal{L}_{img}
\end{equation}
where we set $\lambda_{disc}=1.0$, $\lambda_{aux}=0.01$, and $\lambda_{img}=1.0$ in all of our experiments. The components are defined as follows:
\begin{itemize}
    \item $\mathcal{L}_{discrete}$ (Inherited): The standard cross-entropy loss for next-token prediction. Specifically, this unifies pure language modeling, multimodal understanding, and conditional text prompts for image generation, ensuring model maintains robust instruction following capabilities.
    \item $\mathcal{L}_{aux}$ (Inherited): The auxiliary load-balancing loss is to encourage uniform token distribution, which prevents expert collapse and ensures efficient routing. Crucially, consistent with our architectural disentanglement, this loss is computed independently for the understanding and generation groups. This ensures balanced expert utilization within each modality-specific subspace, rather than enforcing a global balance that might dilute modality specialization.
    \item $\mathcal{L}_{img}$ (New): The flow matching loss on VAE latents, responsible for aligning the visual features with the generative manifold to synthesize high-fidelity images.
\end{itemize}

\section{Experiments}
\label{sec:exp}

\subsection{Experimental Setup}

\noindent\textbf{Experimental Setup.}
All experiments are conducted on a large-scale proprietary corpus spanning T2I, T2I-Long, LM, and MMU tasks. We adopt a holistic evaluation protocol to rigorously assess both modalities.

\smallskip\noindent\textit{Evaluation Metrics.}
Due to space constraints, the main paper focuses on two representative benchmarks for understanding: \textbf{MMLU}~\cite{hendrycks2020measuring} (general knowledge reasoning) and \textbf{OCRBench}~\cite{liu2024ocrbench} (fine-grained visual perception). For generation, we report FID~\cite{heusel2017gans}, CLIPScore~\cite{hessel2021clipscore}, and HPSv2~\cite{wu2023human} on \textbf{COCO-30K}~\cite{lin2014microsoft} to evaluate generation fidelity, alongside \textbf{T2I-CompBench}~\cite{huang2023t2i} for semantic alignment. Extensive evaluations on a broader suite of multimodal understanding and generation benchmarks are provided in 
Supplementary Material Sec. \hyperref[sec:supp_quant]{B}. 
% Supplementary Material Sec. \textcolor{red}{B}.
These extensive results further corroborate the consistent superiority of our method across diverse domains.

\smallskip\noindent\textit{Baselines.}
We compare against two representative architectures under identical data settings: (1) \textbf{Standard MoE}: naive unified co-tuning; and (2) \textbf{MoT}: structural expert partitioning (96/32 split) \textit{without} shared experts. Crucially, all baselines undergo full fine-tuning to isolate the impact of our architectural and training innovations. \textit{Note: Since the generative module is trained from scratch in this pre-training phase, our primary focus is on architectural comparison and training dynamics rather than absolute aesthetic perfection.}

% \smallskip\noindent\textit{Baselines.}
% We compare against three representative architectures under identical data settings: 
% (1)~\textbf{Standard MoE}: naive unified co-tuning; 
% (2)~\textbf{MoT}: structural expert partitioning (96/32 split) \textit{without} shared experts, and uses shared QKVs; and (3)~\textbf{Bagel}~\cite{deng2025bagel}: enforces strict physical isolation (independent QKVs and FFNs). Crucially, all baselines undergo full fine-tuning to isolate the impact of our architectural and training innovations. \textit{Note: Since the generative module is trained from scratch in this pre-training phase, our primary focus is on architectural comparison and training dynamics rather than absolute aesthetic perfection.}

\medskip\noindent\textbf{Implementation Details.}
All models are initialized from the state-of-the-art Hunyuan-A3B (total 30B parameters) VLM. Training is conducted on 256 NVIDIA H20 GPUs with a global batch size of 2500 sample ($\sim$2M tokens) per iteration, optimized via AdamW with 500 warmup steps. The data mixture is fixed at \textit{T2I:T2I-Long:LM:MMU} = $3:3:2:2$. We enforce strict determinism by fixing all random seeds to ensure reproducibility and verified that the setup yields identical training loss curves and evaluation metrics across multiple independent runs. Comprehensive dataset details and hyperparameters are provided in Supplementary Material Sec. \hyperref[sec:supp_impl]{C}.
% Supplementary Material Sec. \textcolor{red}{C}.

\begin{figure}[t] 
    \centering
    % \vspace{-1.5em} 
    \includegraphics[width=1.0\textwidth]{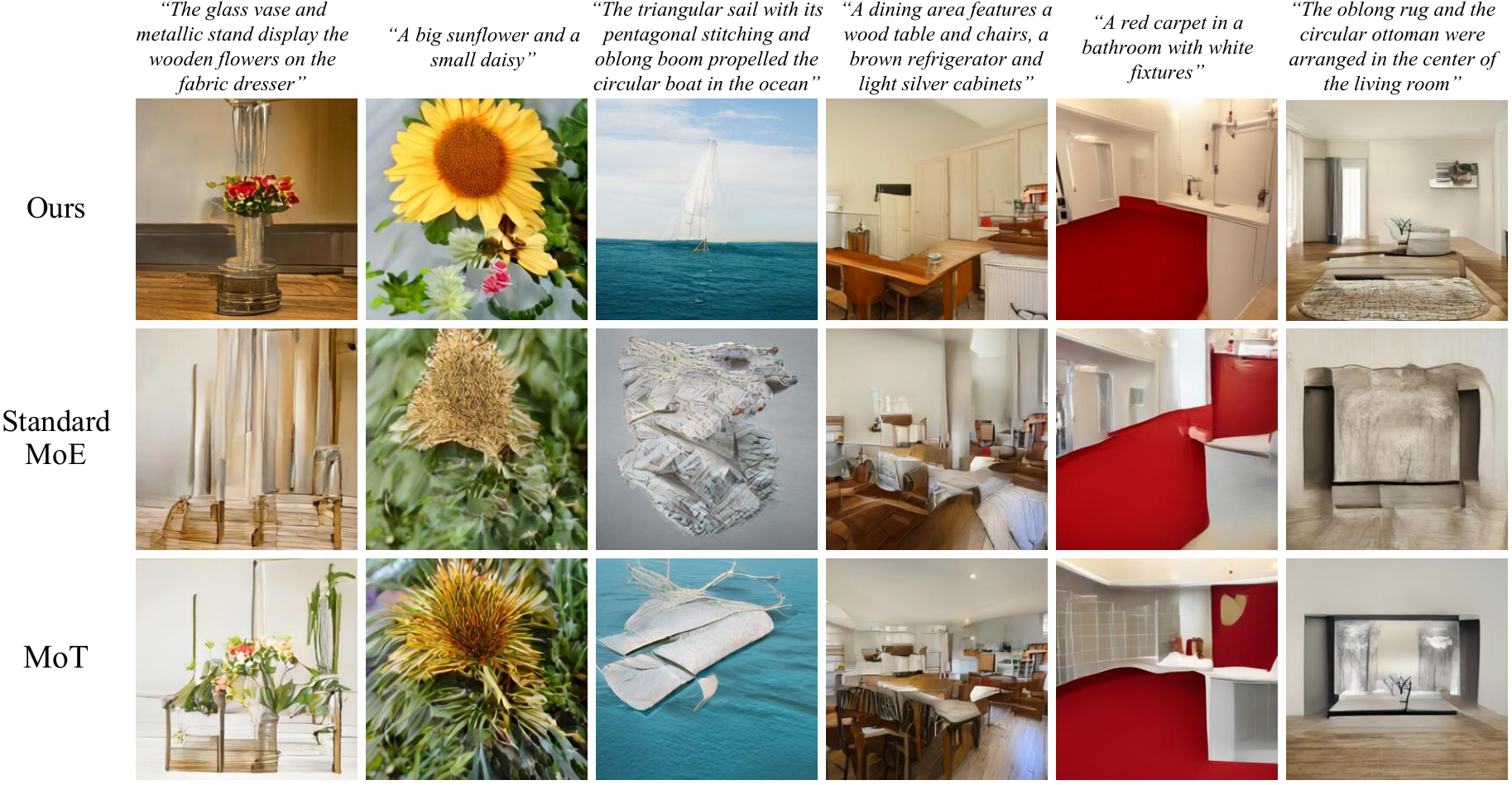}
    \caption{\textbf{{Visualization Comparisons}.} 
    We compare samples generated by Symbiotic-MoE (top), Standard MoE (middle), and MoT (bottom). Standard MoE suffers from severe structural collapse and visual artifacts due to gradient conflicts. While MoT recovers basic object shapes, it often misses fine-grained semantic details. In contrast, our method achieves superior fidelity and precise semantic alignment (\eg, \textit{triangular sail}, \textit{oblong rug}), validating the effectiveness of our symbiotic strategy.
  }
    \label{fig:exp_vis}
    \vspace{-1.0em} 
\end{figure}

\begin{table}[!t]
  \caption{\textbf{Quantitative Results.} 
  We evaluate \textbf{Image Generation} (left) and \textbf{Understanding Capabilities} (right). Standard MoE suffers from severe catastrophic forgetting in understanding tasks. MoT and Bagel (the same as MoT except for using independent QKVs) preserve original capabilities via isolation but lack synergy. Symbiotic-MoE (Ours) delivers superior generation quality while synergistically boosting understanding capabilities (\eg, MMLU, OCRBench) beyond the Only\_LM\_MMU baseline. Best results are highlighted in \textbf{bold}. Additionally, we provide results of Symbiotic-MoE trained for 100k steps (in gray) to showcase its extended scaling capacity.
  }
  \label{tab:quantity}
  \centering
  \resizebox{0.95\textwidth}{!}{
  \begin{tabular}{l|c|ccccccc|cc} 
    \toprule
    \multirow{2.5}{*}{\textbf{Method}} & \multirow{2.5}{*}{\begin{tabular}{@{}c@{}}\textbf{Training} \\ \textbf{Steps}\end{tabular}} & \multicolumn{7}{c|}{\textbf{Image Generation Capabilities}} & \multicolumn{2}{c}{\textbf{Understanding}} \\
    \cmidrule(lr){3-9} \cmidrule(l){10-11} 
    & & T2I-Comp$\uparrow$ & \textit{Color}$\uparrow$ & \textit{Shape}$\uparrow$ & \textit{Texture}$\uparrow$ & FID$\downarrow$ & CLIP$\uparrow$ & HPSv2$\uparrow$ & MMLU$\uparrow$ & OCRBench$\uparrow$ \\
    \midrule
    \textcolor{gray}{Only\_LM\_MMU} & \textcolor{gray}{100k} & \textcolor{gray}{-} & \textcolor{gray}{-} & \textcolor{gray}{-} & \textcolor{gray}{-} & \textcolor{gray}{-} & \textcolor{gray}{-} & \textcolor{gray}{-} & \textcolor{gray}{0.405} & \textcolor{gray}{662} \\
    \midrule
    Standard MoE & 100k & 0.36 & 0.38 & 0.21 & 0.40 & 26.27 & 0.27 & 0.19 & 0.308 & 571 \\
    MoT & 100k & 0.43 & 0.51 & 0.27 & 0.48 & 19.87 & 0.28 & 0.21 & 0.392 & 583 \\
    Bagel & 100k & 0.45 & 0.52 & 0.29 & 0.51 & 18.42 & 0.29 & 0.21 & 0.396 & 590 \\
    \textbf{Symbiotic-MoE} & 100k & \textbf{0.49} & \textbf{0.55} & \textbf{0.35} & \textbf{0.56} & \textbf{13.65} & \textbf{0.31} & \textbf{0.23} & \textbf{0.507} & \textbf{768} \\
    \bottomrule
  \end{tabular}
  }
  \vspace{-1em} 
\end{table}

\subsection{Main Results}

\noindent\textbf{Image Generation.} 
We evaluate generative fidelity and semantic alignment on \textit{COCO-30K} (FID, CLIP, HPSv2) and \textit{T2I-CompBench (color, shape, texture)}. As detailed in Table~\ref{tab:quantity}, Standard MoE suffers from routing collapse, resulting in severe artifacts and a high FID of 26.37. While MoT and Bagel~\cite{deng2025bagel} improve stability via isolation, it lacks the semantic guidance required for complex prompts, resulting 0.43 and 0.45 respectively on T2I-Compbench~\cite{huang2023t2i}. In contrast, our Symbiotic-MoE achieves best performance across all metrics. 
We ascribe this superior fidelity to the \textit{holistic synergy} of our architecture: the specialized \textit{Generation Group} effectively captures high-frequency visual details, while the \textit{Shared Expert bridge} anchors these details to robust semantic representations. 
Qualitative comparisons in Fig.~\ref{fig:exp_vis} corroborate this. As highlighted in the 3\textsuperscript{rd} column, our model accurately synthesizes geometrically complex objects like the ``\textit{triangular sail}'', whereas baselines struggle with shape consistency. Similarly, in the 6\textsuperscript{th} column, Symbiotic-MoE successfully disentangles spatial relationships (placing the ``\textit{oblong rug}'' correctly), avoiding the structural distortion observed in MoT. Additional qualitative results are provided in Supplementary Material Sec. \hyperref[sec:supp_vis]{D}.
% Supplementary Material Sec. \textcolor{red}{D}.

\medskip\noindent\textbf{Understanding Capabilities.}
To verify the understanding ability, we report results on MMLU~\cite{hendrycks2020measuring} (reasoning) and OCRBench~\cite{liu2024ocrbench} (perception) in Table~\ref{tab:quantity}.
Standard MoE succumbs to severe gradient conflict, suffering catastrophic forgetting with MMLU scores plummeting to 0.308. While MoT and Bagel mitigate this decay via physical isolation, they merely maintain a baseline level of $0.392$ and $0.396$ respectively, failing to leverage the visual signals from the generative stream. Crucially, Symbiotic-MoE achieves a breakthrough. 
% It not only outperforms both standard MoE and MoT baselines by a significant margin but, most notably, surpasses the \textit{Only\_LM\_MMU} control group—which was trained exclusively on understanding tasks without generative interference using the same setting of ours. 
It not only outperforms both standard MoE, MoT, and Bagel baselines by a significant margin but, most notably, surpasses \textit{Only\_LM\_MMU}—which was trained identically to Symbiotic-MoE but without T2I to isolate the data domain shifts that inherently degrade the VLM.
Specifically, our method boosts MMLU performance from $0.405$ to $0.507$ (+20.1\% relative gain) and OCRBench from $662$ to $768$ (+13.8\%). 
This empirical evidence challenges the prevailing view that generation and understanding are zero-sum competitors. Instead, it validates our core hypothesis: pixel-level generative training, when properly orchestrated via our shared-expert bridge, acts as a powerful fine-grained visual regularizer. It forces the visual encoder to capture denser semantic details required for reconstruction, which reciprocally enhances the model's discriminative reasoning capabilities.

\begin{figure}[t]
  \centering
  \begin{subfigure}[b]{0.40\linewidth}
    \centering
    \includegraphics[width=\linewidth]{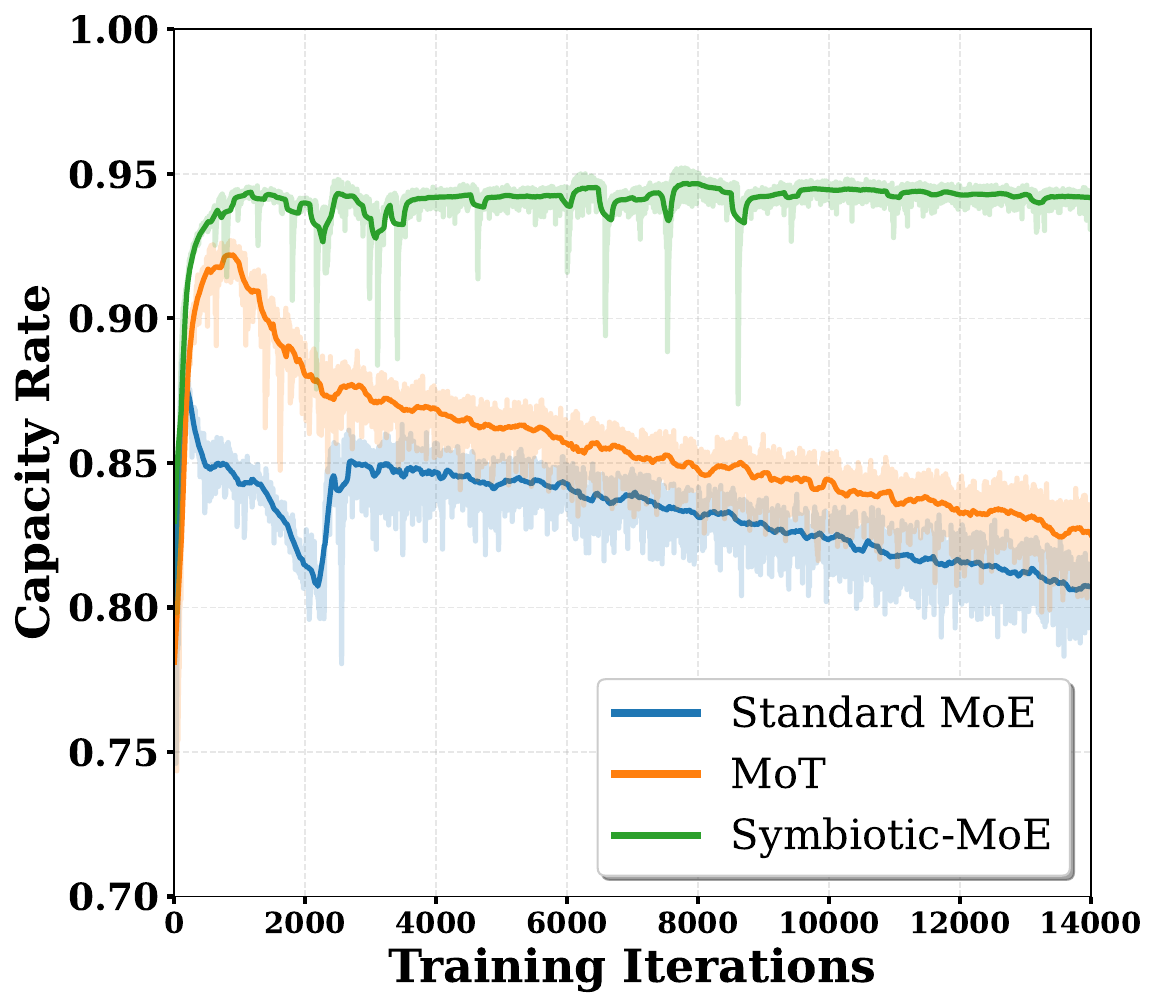} 
    \caption{Expert Capacity Rate}
    \label{fig:capacity}
  \end{subfigure}
  \hspace{0.5em}
  \begin{subfigure}[b]{0.42\linewidth}
    \centering
    \includegraphics[width=\linewidth]{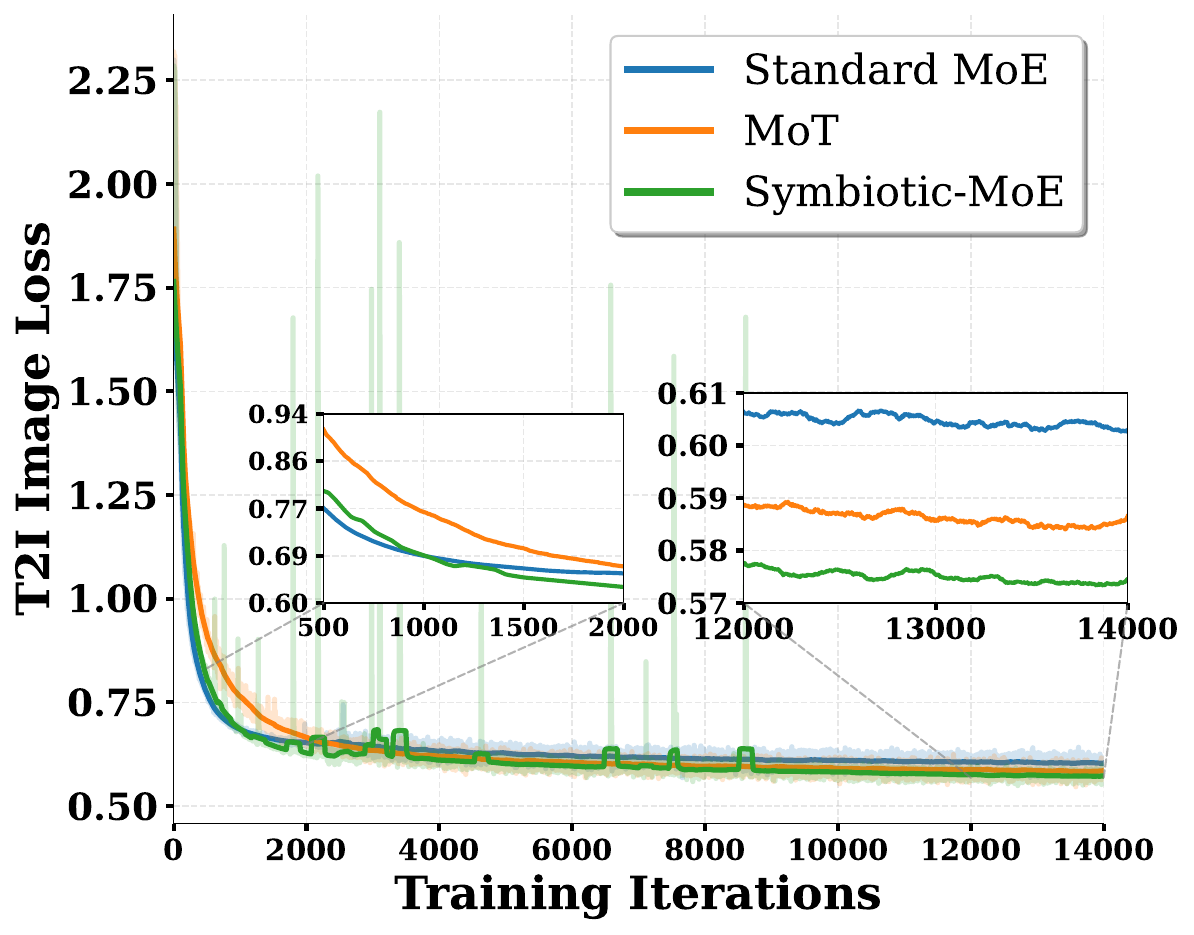}
    \caption{T2I Training Loss}
    \label{fig:t2i_loss}
  \end{subfigure}
  
  \vspace{-5pt}
  \caption{
    \textbf{Analysis of Training Dynamics.} 
    \textbf{(a)} Standard MoE (blue) and MoT (orange) suffer from routing collapse, indicated by the dropping capacity rate (ratio of non-dropped tokens). In contrast, our Symbiotic-MoE (green) maintains highest expert utilization ($\sim$0.95).
    \textbf{(b)} This structural stability translates into superior optimization efficiency, with our method achieving lower convergence generation loss than baselines.
  }
  \label{fig:training_dynamics}
  \vspace{-10pt}
\end{figure}

\medskip\noindent\textbf{Training Efficiency.}
Beyond final performance, we analyze the training dynamics to uncover the source of our model's efficiency. As illustrated in Fig.~\ref{fig:capacity}, standard MoE suffers from severe routing collapse, where expert utilization (capacity rate) drops significantly. In contrast, Symbiotic-MoE maintains an exceptionally high capacity rate, approaching \textbf{0.95}, a value that significantly exceeds the typical equilibrium threshold of 0.90. This confirms that our method effectively enforces a near-perfect load balance across experts. 

Crucially, this structural efficiency translates directly into better optimization. Figure~\ref{fig:t2i_loss} Text-to-image Training Loss reveals a turning point around 1,000 iterations, where our method surpasses the baseline MoE in convergence speed and consistently maintains a lower loss thereafter. This provides strong empirical evidence that \textit{a compact set of 32 specialized generative experts, supported by a shared expert, is more effective than a massive pool of 128 entangled generalists plus one shared expert.} This finding underscores that for multimodal MoE, architectural specialization via the allocation of specific experts to specific modalities is more critical than simply scaling the total number of available experts.

\begin{table}[t]
    \centering
    \caption{\textbf{Ablation on Architectural Designs.} We evaluate knowledge retention at iteration zero to validate structural integrity. ``Tripartite'' denotes separating Text, ViT, and VAE; ``Bimodal'' merges Text/ViT into a single group. The 96/32 Bimodal split with shared experts offers the optimal trade-off.}
    \label{tab:ablation_grouping}
    \setlength{\tabcolsep}{10pt} 
    \resizebox{0.95\linewidth}{!}{
    \begin{tabular}{l|ccc|c|cc}
        \toprule
        \multirow{2}{*}{\textbf{Grouping Strategy}} & \multicolumn{3}{c|}{\textbf{Expert Allocation (Count)}} & \textbf{Shared} & \multicolumn{2}{c}{\textbf{Zero-Shot Metrics}} \\
        \cmidrule{2-4} \cmidrule{6-7}
         & \textbf{Text} & \textbf{ViT} & \textbf{VAE} & \textbf{Expert} & \textbf{MMLU} & \textbf{OCRBench} \\
        \midrule
        \rowcolor{gray!10} Original VLM & \multicolumn{2}{c}{128 (Entangled)} & - & $\checkmark$ & 0.697 & 845 \\
        \rowcolor{gray!10} -- w/o Shared Expert & \multicolumn{2}{c}{128 (Entangled)} & - & $\times$ & 0.460 & 498 \\
        \midrule
        \multirow{2}{*}{Tripartite Split} & 32 & 32 & 64 & $\checkmark$ & 0.244 & 109 \\
         & 44 & 42 & 42 & $\checkmark$ & 0.234 & 179 \\
        \midrule
        \multirow{4}{*}{Bimodal Split} & \multicolumn{2}{c}{32 (Understanding)} & 96 & $\checkmark$ & 0.285 & 532 \\
         & \multicolumn{2}{c}{64 (Understanding)} & 64 & $\checkmark$ & 0.453 & 742 \\
         & \multicolumn{2}{c}{86 (Understanding)} & 42 & $\checkmark$ & 0.561 & 786 \\
         & \multicolumn{2}{c}{\textbf{96 (Understanding)}} & \textbf{32} & $\checkmark$ & \textbf{0.601} & \textbf{807} \\
        \bottomrule
    \end{tabular}
    }
\vspace{-1em}
\end{table}

\subsection{Ablation Studies}
\noindent\textbf{Architectural Design Analysis.} 
We first investigate the impact of expert partitioning and shared experts at iteration zero on knowledge retention (Table~\ref{tab:ablation_grouping}). 

\medskip\noindent\textit{Expert Disentanglement Strategy.} 
We first hypothesize a granular \textit{Tripartite Split} (separating Text, ViT, and VAE). However, this configuration triggers a catastrophic collapse (MMLU $0.69 \!\to\! 0.24$), revealing a critical functional overlap between Text and ViT experts in the original VLM. Forcibly separating them severs essential semantic pathways. Consequently, we adopt a \textit{Bimodal Split}, consolidating Text and ViT into a unified Understanding Group. By evaluating different assignment ratios, we identify the 96/32 split as the optimal configuration. This setting maximizes understanding retention ($0.60$) while reserving sufficient plasticity (32 experts) for the generative task.

\medskip\noindent\textit{Necessity of Shared Expert Bridge.} 
Table~\ref{tab:ablation_grouping} (row 2) further validates the role of shared experts. Removing them from the original VLM precipitates a sharp performance decline ($0.69 \!\to\! 0.46$). This confirms that shared experts act as a non-negotiable semantic anchor, ensuring feature alignment across specialized expert groups.

\begin{figure}[t]
    \centering
    \begin{subfigure}{0.34\textwidth}
        \includegraphics[width=\linewidth]{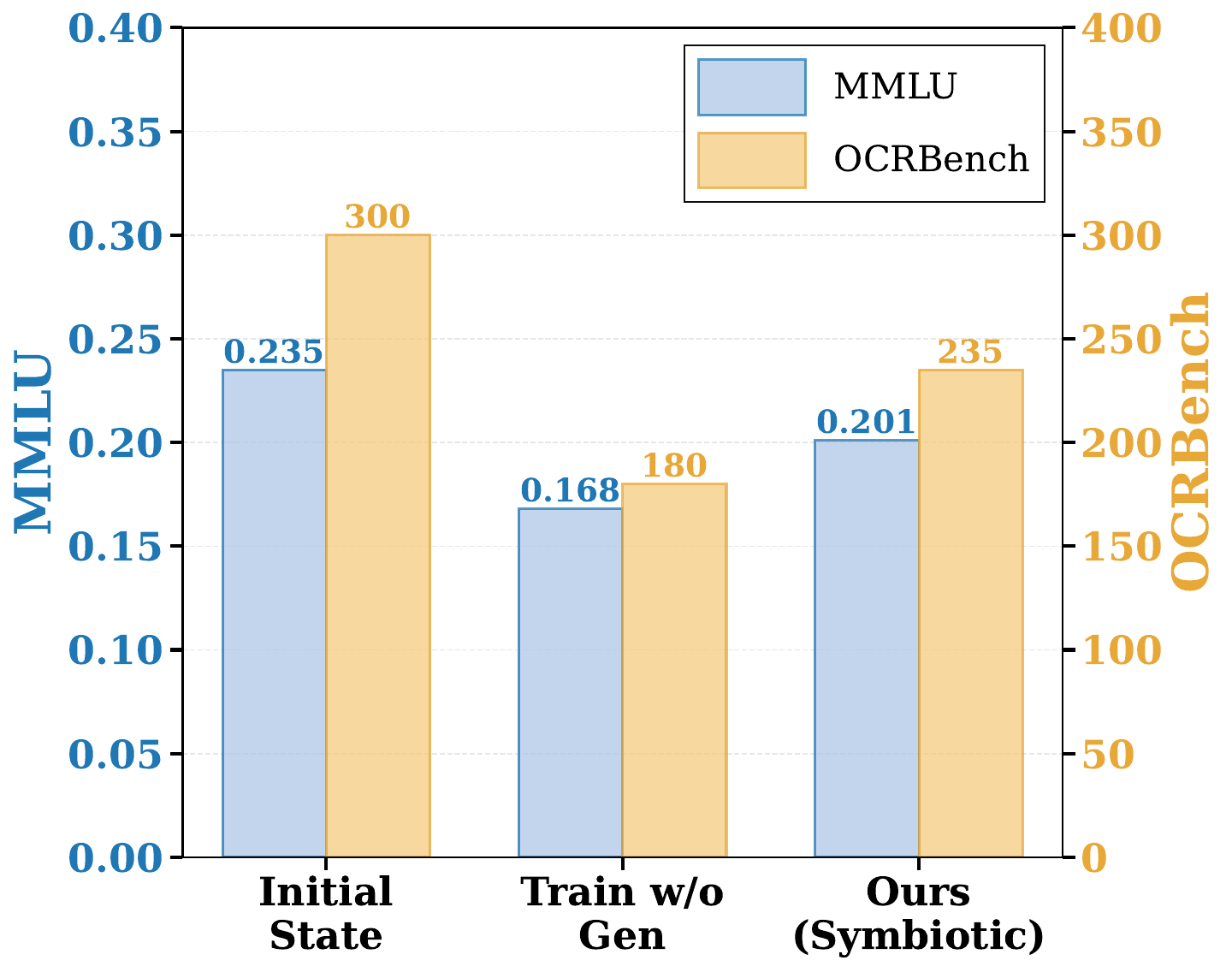} 
        \caption{Shared Expert Probing}
        \label{fig:synergy_probing}
    \end{subfigure}
    \hfill
    \begin{subfigure}{0.32\textwidth}
        \includegraphics[width=\linewidth]{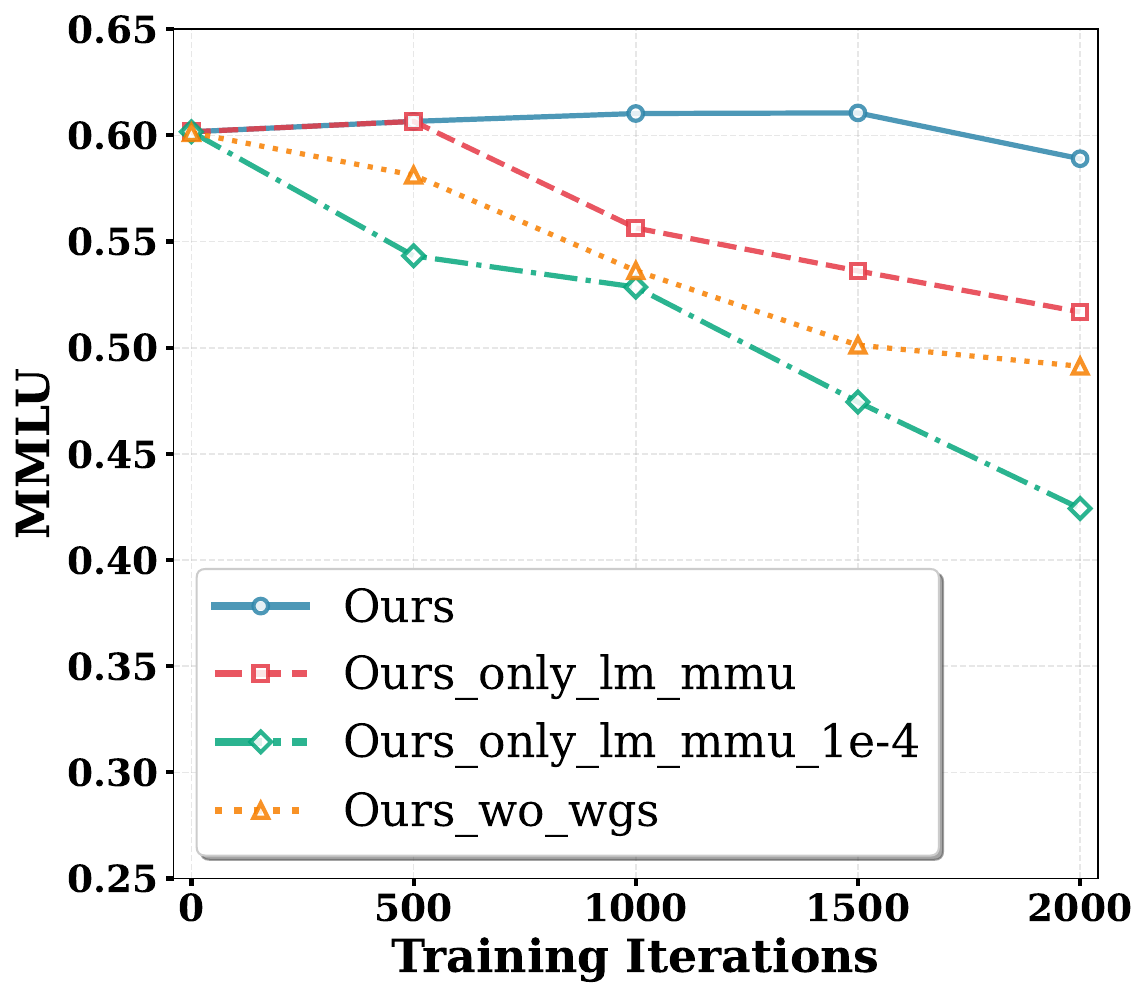}
        \caption{MMLU Dynamics}
        \label{fig:synergy_mmlu}
    \end{subfigure}
    \hfill
    \begin{subfigure}{0.32\textwidth}
        \includegraphics[width=\linewidth]{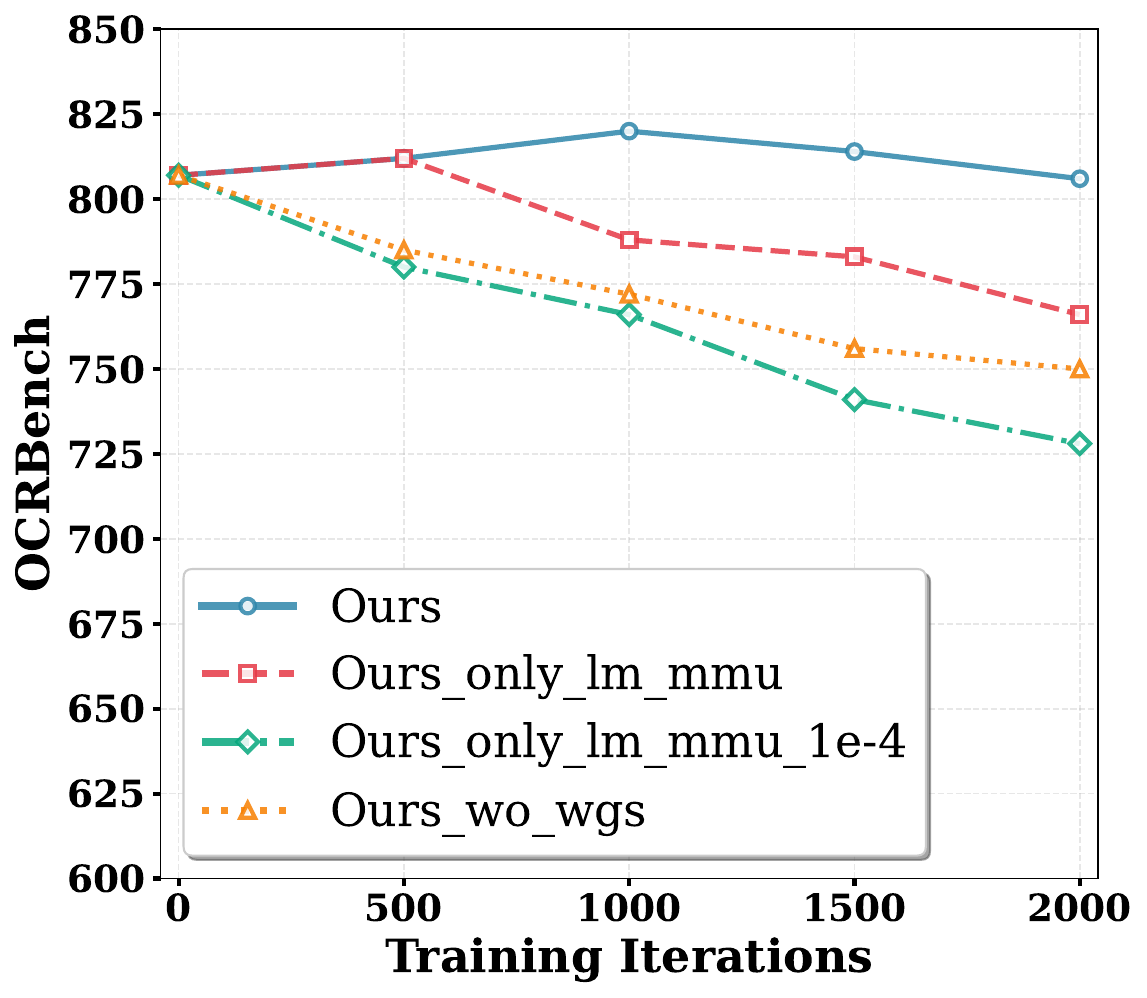}
        \caption{OCRBench Dynamics}
        \label{fig:synergy_ocr}
    \end{subfigure}
    \caption{\textbf{Evidencing Generative Synergy.} (a) We probe the isolated performance of Shared Experts. Compared to the baseline trained without generation (\textit{Train w/o Gen}), our Symbiotic method significantly boosts the semantic density of shared experts. (b) \& (c) The training dynamics show that while the control baseline degrades, Symbiotic-MoE effectively reverses the forgetting trend via generative regularization.
    }
    \label{fig:synergy}
    \vspace{-2em}
\end{figure}

\medskip\noindent\textbf{Optimization Dynamics Analysis.} 
We decouple the effects of two critical strategies designed to navigate the stability-plasticity dilemma:

\medskip\noindent\textit{Differential Learning Rates.} 
We identify a critical optimization mismatch: generative learning demands aggressive updates, whereas the pre-trained backbone requires conservative fine-tuning. Crucially, as shown by the green dashed line (\texttt{Ours\_only\_lm\_mmu\_1e-4}) in Fig.~\ref{fig:synergy}(b--c), training \textit{solely} on understanding tasks at $1e^{-4}$ triggers immediate collapse even without generative interference. This exposes a fundamental insight: catastrophic forgetting here is driven less by task conflict and more by the backbone's intrinsic intolerance to high-magnitude updates. Consequently, we enforce a hierarchical strategy, assigning $1e^{-4}$ to generative experts for plasticity, while restricting shared/understanding experts to a conservative $1e^{-6}$ to respect their stability constraints.

\medskip\noindent\textit{Warmup Gradient Shielding.} 
Generative modules trained from scratch inherently emit high-variance gradients during early optimization. Directly exposing shared experts, the model's semantic anchor, to this volatility causes an ``initial shock'', washing out pre-trained representations before generative features become semantically meaningful. To mitigate this, we explicitly detach the generative gradient flow to shared experts during warmup. As evidenced by the orange dashed line (\texttt{Ours\_wo\_wgs})  in Fig.~\ref{fig:synergy}(b--c), removing this shield results in a precipitous decline in MMLU and OCRBench during the first 500 iterations, confirming that early stage isolation is vital for stability.

\subsection{Analysis of Synergy}
\label{sec:synergy}

While standard MoE training suffers from interference, Symbiotic-MoE orchestrates a constructive interference, synergy. To rigorously verify this, we deconstruct the interaction between modalities through three analytical lenses: component probing, subtractive ablation, and optimization dynamics.

\medskip\noindent\textbf{Component Isolation of Shared Experts.} 
To pinpoint the architectural locus of cross-modal synergy, we probe the \textit{Shared Experts} in isolation. By applying a hard mask to all modality-specific routed experts, we enforce zero-shot inference using only the shared expert parameters. As visualized in Fig.~\ref{fig:synergy}(a), the baseline (\texttt{Train w/o Gen}) optimized exclusively on understanding tasks ($\lambda_{img}=0$) suffers from representational degradation in the shared module. In contrast, the shared experts within Symbiotic-MoE demonstrate a substantial performance resurgence, consistently outperforming the understanding-only baseline on both MMLU and OCRBench by a clear margin. This empirical evidence validates that generative gradients do not overwrite linguistic priors, rather, they function as a dense semantic compressor, forcing the shared parameters to encapsulate highly generalized representations.

\medskip\noindent\textbf{Generative Regularization: Does Painting Help Seeing?}
We isolate the impact of generative training via a counter-factual baseline trained \textit{solely} on understanding tasks. As visualized in Fig.~\ref{fig:synergy_mmlu}, the baseline (red dashed line) suffers from severe knowledge decay (0.60 $\to$ 0.52 on MMLU) driven by distribution shift. Crucially, incorporating generation arrests this decline. Our Symbiotic-MoE (blue solid line) not only stabilizes general reasoning but significantly boosts fine-grained perception, propelling OCRBench to a peak of 820 (\vs 790). We attribute this to the unique nature of generative objectives: unlike discriminative tasks that often allow for semantic shortcuts, pixel-level reconstruction compels the shared experts to capture precise spatial relationships. This strictly constrains the representation space, effectively acting as a regularizer that prevents overfitting to textual patterns and reciprocally sharpens visual perception.

\medskip\noindent\textbf{Acceleration in Optimization.} 
Finally, we examine if strong understanding capabilities benefit generation. Figure~\ref{fig:t2i_loss} illustrates that the T2I convergence of Symbiotic-MoE is significantly faster and deeper than the Standard MoE and MoT baselines. This bi-directional flexibility aligns the semantic space with the generative manifold, turning potential conflicts into a driver for comprehensive capability emergence.

\section{Conclusion}
\label{sec:conclusion}

In this work, we presented Symbiotic-MoE, a unified pre-training framework that successfully reconciles the long-standing catastrophic forgetting in extending VLMs with generative capabilities. By moving beyond the structural isolation typical of prior arts like MoT, we demonstrated that task interference is not an intrinsic limitation of unified architectures, but rather an optimization challenge resolvable through modality-aware disentanglement and progressive training dynamics. Most significantly, our empirical results challenge the conventional zero-sum view of multimodal training. We show that when the routing landscape is carefully orchestrated, the fine-grained visual semantics acquired from generation can retroactively refine the model's understanding capability, rather than eroding them. We hope this work serves as a blueprint for future native omni-modal foundation models, establishing that the path to true general intelligence lies not in the segregation of perception and creation, but in their symbiotic integration within a single, efficient parameter space.
\clearpage
\appendix % 开始附录模式，章节编号变为 A, B, C...

% =======================================================
% 1. 计数器重置设置 
% =======================================================
\renewcommand{\thesection}{\Alph{section}}
\renewcommand{\thefigure}{\arabic{figure}}
\renewcommand{\thetable}{\arabic{table}}
\renewcommand{\theequation}{\arabic{equation}}

% =======================================================
% 2. 【核心修复】
% =======================================================
% 这几行专门给 hyperref 引路，避免内部 ID 和正文重名
\renewcommand{\theHsection}{Supp\Alph{section}}
\renewcommand{\theHfigure}{Supp\arabic{figure}}
\renewcommand{\theHtable}{Supp\arabic{table}}
\renewcommand{\theHequation}{Supp\arabic{equation}}

\title{Supplementary Materials for\texorpdfstring{\\}{ }Symbiotic-MoE: Unlocking the Synergy between Generation and Understanding}
\titlerunning{Symbiotic-MoE Supplementary Material}
\author{\vspace{-5ex}} 
\institute{\vspace{-5ex}}
\maketitle

% --- Overview ---
\noindent This supplementary material complements the main paper by providing in-depth architectural analyses, extended evaluation results, and detailed reproduction specifications. The content is organized as follows:
\begin{itemize}[leftmargin=*]
    \item \textbf{Section \ref{sec:supp_arch}}: In-depth analysis of expert specialization, providing comprehensive empirical justification for our modality-aware grouping strategy.
    \item \textbf{Section \ref{sec:supp_quant}}: Additional quantitative evaluations on diverse benchmarks and fine-grained training dynamics analysis.
    \item \textbf{Section \ref{sec:supp_impl}}: Detailed hyperparameters, dataset composition, and computational costs.
    \item \textbf{Section \ref{sec:supp_vis}}: Additional qualitative results for image generation.
\end{itemize}

% =======================================================
% Section A: Architecture Analysis 
% =======================================================
\section{In-Depth Architectural Analysis}
\label{sec:supp_arch}

\subsection{Expert Specialization Analysis: The Rationale for Bimodal Split}
\label{sec:supp_expert_analysis}

In the main paper (Sec.~\ref{sec:architecture}), we proposed \textit{Modality-Aware Expert Disentanglement}, partitioning the experts into a unified Understanding Group (Text+ViT) and a Generation Group (VAE). To validate the empirical foundation of this design, we conduct an in-depth visualization of the intrinsic activation patterns within the pre-trained VLM prior to any fine-tuning. 

\begin{figure}[t]
  \centering 
  \begin{subfigure}[b]{0.40\linewidth}
    \centering
    \includegraphics[width=\linewidth]{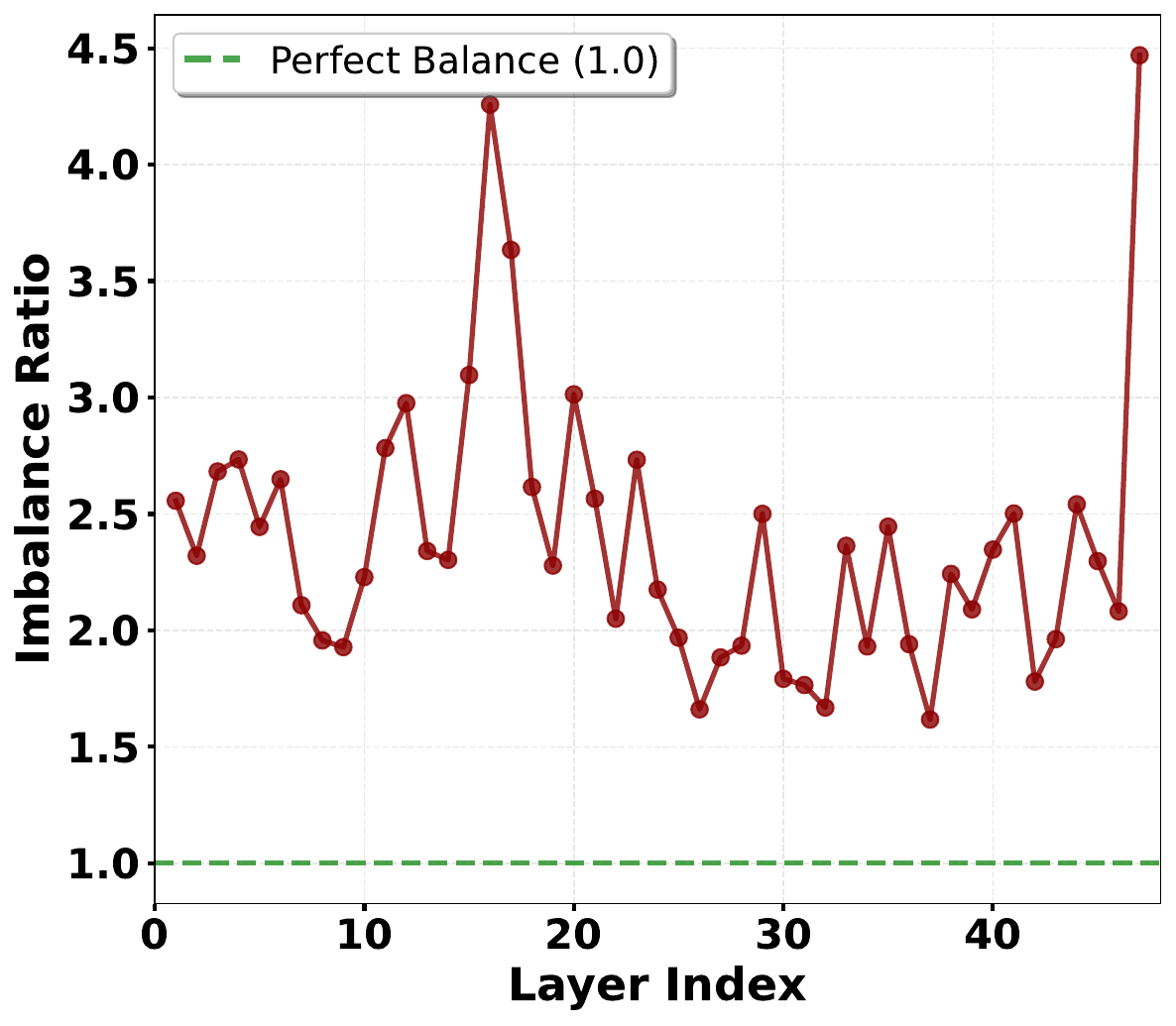} 
    \caption{Imbalance Ratio}
    \label{fig:layer_imbalance_ratio}
  \end{subfigure}
  \hspace{0.5em} 
  \begin{subfigure}[b]{0.40\linewidth}
    \centering
    \includegraphics[width=\linewidth]{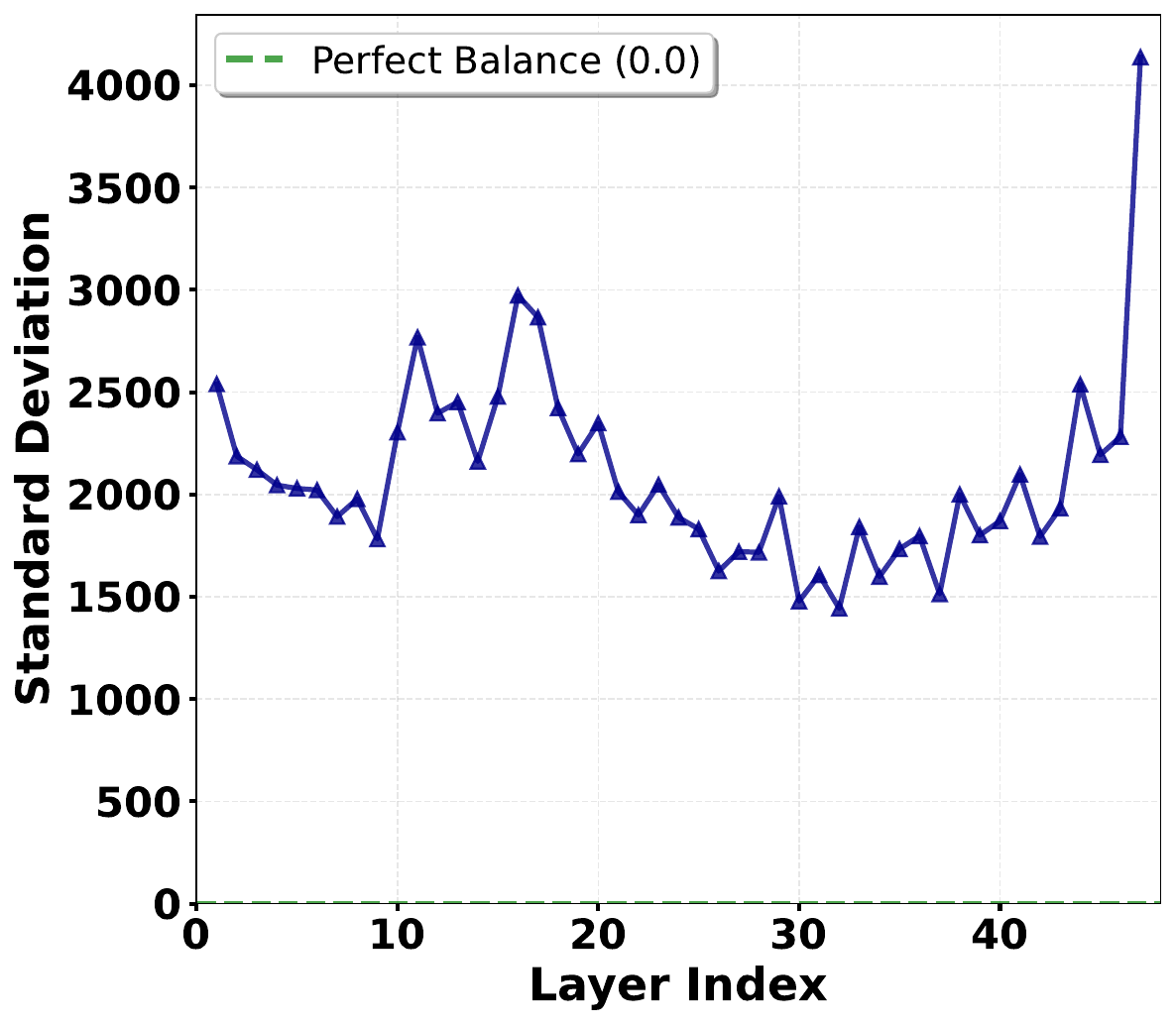}
    \caption{Standard Deviation}
    \label{fig:layer_standard_deviation}
  \end{subfigure}
  
  \vspace{-5pt}
  \caption{
    \textbf{Macro-level routing dynamics of the pre-trained VLM.} We track the (a) imbalance ratio and (b) standard deviation of expert selection counts across all 47 layers for Text tokens. The metrics reveal a severe long-tail distribution (ratio consistently $\gg 1.0$), indicating that the pre-trained VLM inherently relies on a concentrated subset of ``core experts'' while leaving others underutilized.
  }
  \label{fig:expert_activation}
  \vspace{-10pt}
\end{figure}

\medskip\noindent\textbf{Macro-Level Routing Dynamics.} 
To understand the internal allocation of model capacity, we first investigate the global routing distribution across all 47 MoE layers. Specifically, we establish two metrics to quantify the activation frequencies of Text tokens (using MMLU~\cite{hendrycks2020measuring}) and ViT tokens (using OCRBench~\cite{liu2024ocrbench}):
(1) \textbf{Imbalance Ratio}: Defined as the ratio of the maximum expert selection count to the mean selection count ($\frac{\max}{\text{mean}}$). An ideal uniform routing would yield a ratio of 1.0. An ideal uniform routing would yield a ratio of 1.0, whereas higher values indicate a high concentration of routing density onto a few ``popular'' experts.  
(2) \textbf{Standard Deviation (Std)}: Measures the absolute dispersion of selection counts from the mean. 
As illustrated in Fig.~\ref{fig:expert_activation}, the routing mechanism exhibits a heavily skewed, long-tail distribution. For instance, the Imbalance Ratio frequently exceeds 2.5 and reaches up to 4.26 (\eg, in Layer 16). This pronounced imbalance reveals that the original VLM does not utilize experts uniformly; instead, a specific subset of ``core experts'' dominates the processing.

\begin{figure}[t!]
  \centering 
  \begin{subfigure}[b]{0.90\linewidth}
    \centering
    \includegraphics[width=\linewidth]{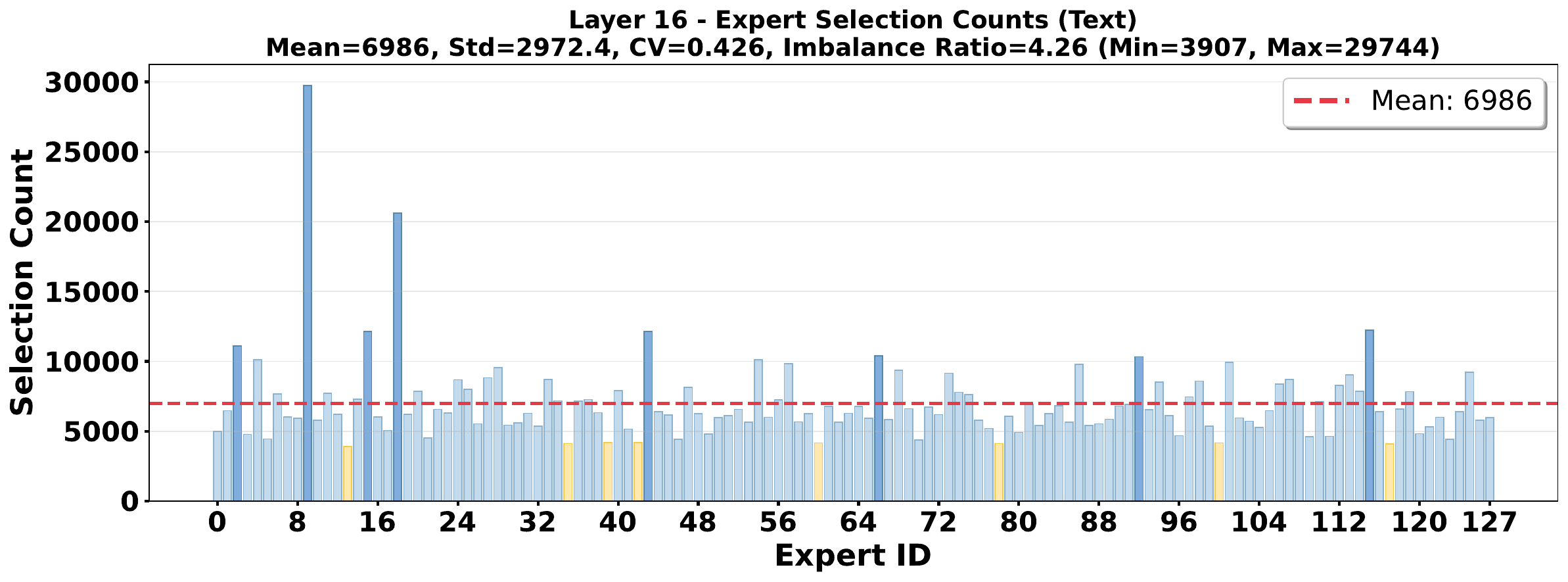} 
    \label{fig:layer_16_text}
  \end{subfigure}
  \vspace{-1.2em} 
  \begin{subfigure}[b]{0.90\linewidth}
    \centering
    \includegraphics[width=\linewidth]{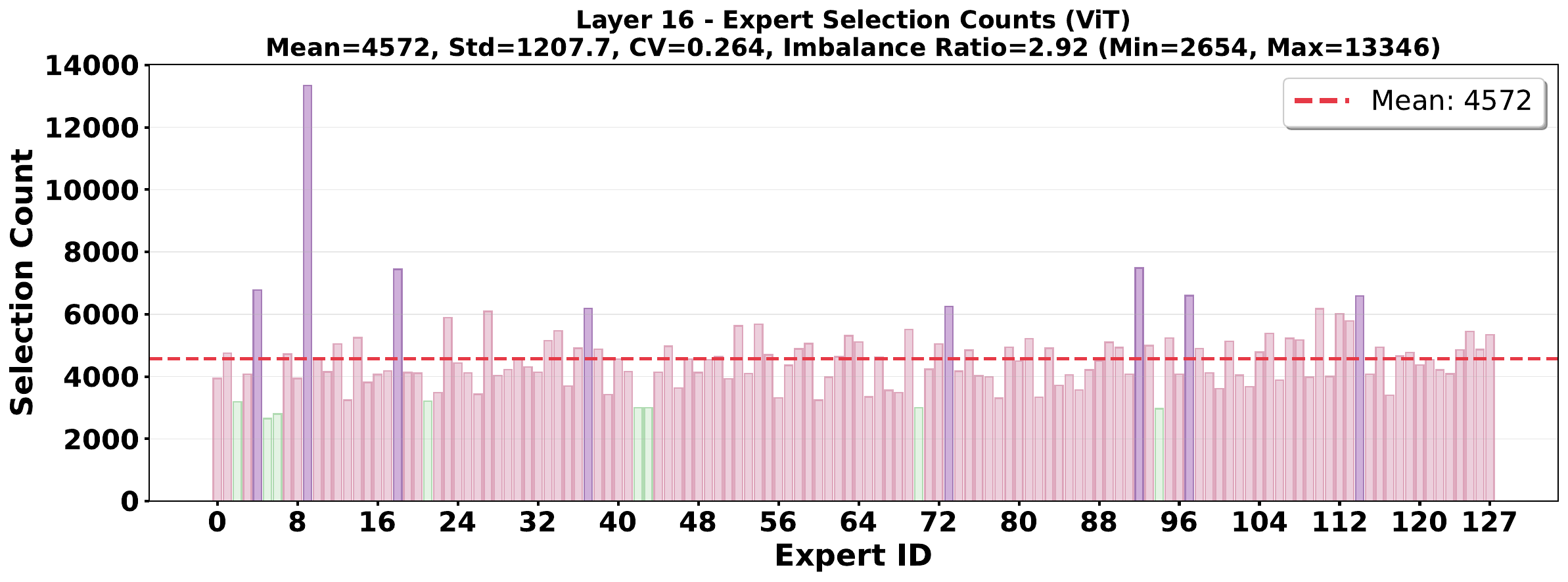}
    \label{fig:layer_16_vit}
  \end{subfigure}
  \vspace{-8pt}
  \caption{
    \textbf{Micro-level modality entanglement in Layer 16.} Activation frequencies of all 128 experts for Text (top) and ViT (bottom) tokens. The top-8 most activated experts are highlighted in dark colors (blue/purple), and the bottom-8 in light colors (yellow/green). A profound intersection is observed (\eg, Experts 16, 52, and 122 are heavily utilized by both modalities in top-8 experts), demonstrating that linguistic and visual perception pathways are intrinsically coupled within the same core experts in the pre-trained VLM.
  }
  \label{fig:expert_layer_16}
  \vspace{-20pt}
\end{figure}

\medskip\noindent\textbf{Micro-Level Modality Entanglement.} 
Given that a small subset of experts dominates, the critical question is: \textit{Do Text and ViT tokens rely on different core experts, or do they share the same ones?} To answer this, we zoom into the granular activation profiles of individual layers. 

Figure~\ref{fig:expert_layer_16} visualizes the selection counts for Text and ViT tokens in Layer 16, which serves as an example for the micro-level routing behavior. We highlight the top-8 most frequently activated experts (dark blue for Text, dark purple for ViT) and the bottom-8 least activated experts (yellow/green). A striking phenomenon emerges: \textbf{there is a profound intersection between the most heavily utilized experts for Text and ViT modalities in the original VLM}. For instance, Expert IDs 9, 18, and 92 are among the top choices for both modalities in top-8 experts. This high degree of coupling is not an isolated artifact but a consistent structural trait observed across the network's depth.

\medskip\noindent\textbf{Conclusion: Why a Bimodal Split?} 
These probing results provide compelling empirical evidence for our architectural decision. The original VLM has intrinsically coupled linguistic and visual perception pathways into a shared subset of "generalist" experts. Forcibly severing these modalities into a granular tripartite split (\ie, isolated Text, ViT, and VAE groups) would violently rupture these learned semantic connections, explaining the catastrophic performance drop observed in our ablation studies in the main paper Table~\ref{tab:ablation_grouping}. Consequently, consolidating Text and ViT into a unified \textit{Understanding Group} while separating the VAE tokens emerges as the optimal strategy to balance pre-trained stability with generative plasticity.

\subsection{Empirical Visualization of Routing Collapse}
\label{sec:supp_conflicts}

To validate the gradient conflicts that trigger the stability-plasticity dilemma in standard MoE, we visualize the routing probabilities of text tokens across the 48 layers (Y-axis) and 128 experts (X-axis) of a standard MoE baseline before and after introducing the text-to-image (T2I) objective. 

\begin{figure}[ht]
    \centering
    \includegraphics[width=0.90\textwidth]{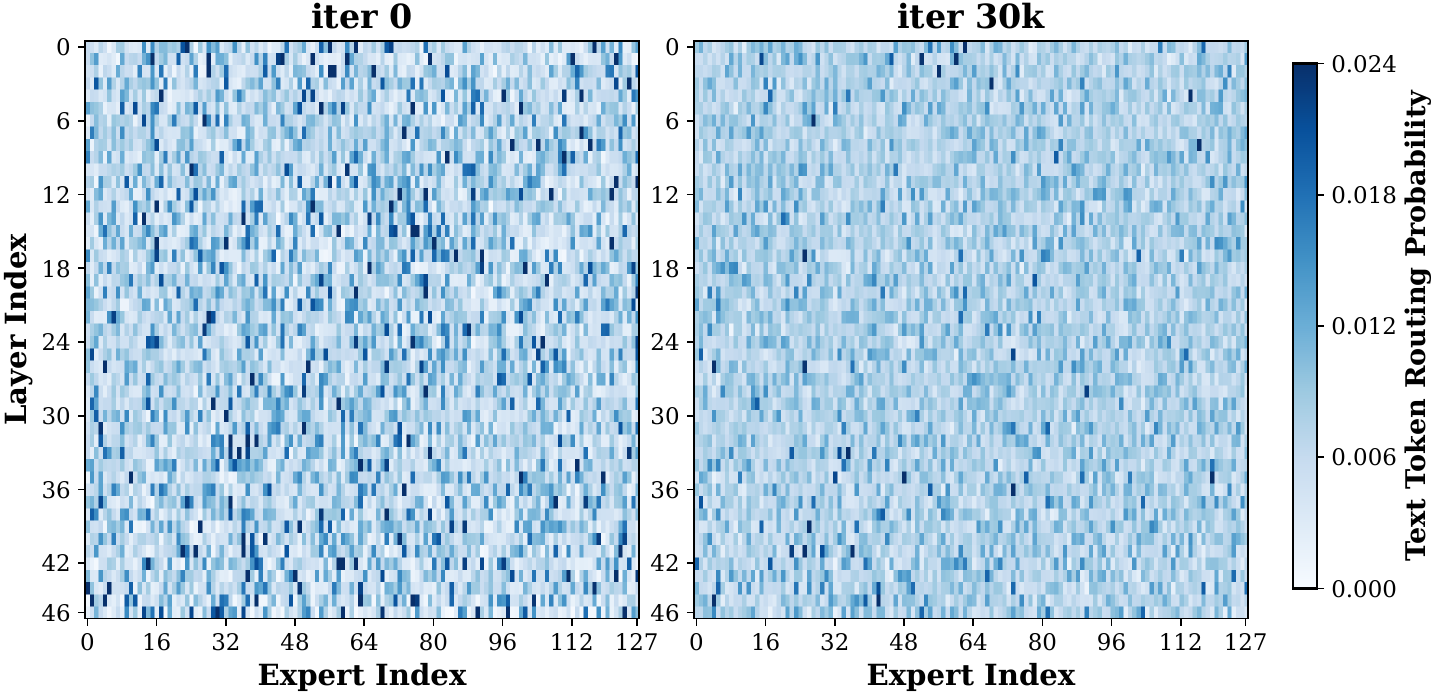}
    \caption{\textbf{Text routing probability in standard MoE.} Left (\textit{iter 0}): Text routing probability before T2I training. Right (\textit{iter 30k}): Text routing probability after naive co-training with T2I, showing that generative gradients aggressively overwrite established language pathways.}
    \label{fig:routing_collapse}
\end{figure}

As shown in Fig.~\ref{fig:routing_collapse}, at initialization (\textit{iter 0}), the text tokens demonstrate highly specialized, sparse, and sharp routing pathways (distinct dark-blue spots), indicating that the router has successfully preserved its pretrained selectiveness. However, after naive co-training with the T2I objective (\textit{iter 30k}), these structured pathways are almost entirely erased, leaving a smeared, blurred, and homogenized routing landscape. This severe routing collapse physically evidences that the aggressive, high-magnitude generative gradients aggressively overwrite established language pathways, leading to uncontrollable expert drift and catastrophic forgetting, thereby validating the co-training dilemma.

\section{Extended Quantitative Evaluation and Dynamics}
\label{sec:supp_quant}

In this section, we provide exhaustive quantitative results deferred from the main manuscript due to space limitations, alongside a fine-grained analysis of the internal routing dynamics to validate the structural health of our Symbiotic-MoE.

\subsection{Comprehensive Benchmark Results}
\label{subsec:full_benchmarks}

To rigorously ensure that the integration of generative objectives introduces no implicit degradation across specialized understanding domains, we expand our evaluation beyond MMLU~\cite{hendrycks2020measuring} and OCRBench~\cite{liu2024ocrbench} to a comprehensive suite of multimodal benchmarks. To guarantee a standardized, reproducible, and unbiased assessment, all evaluations are systematically conducted utilizing the open-source VLMEvalKit~\cite{duan2024vlmevalkit} toolkit. Rather than reducing these diverse metrics to a monolithic score, we decouple them into distinct functional dimensions, providing a granular and multifaceted analysis of the model's capabilities (detailed in Table~\ref{tab:supp_understanding}):

\medskip\noindent\textit{Text Scene Recognition} (\eg, TextVQA~\cite{kembhavi2017you}): Building upon the robust OCR capabilities shown in the main text, these tasks require precise grounding and interpretation of text within natural scenes. We observe that the fine-grained, pixel-level supervision derived from the generative task acts as a strong visual regularizer. Via the shared expert bridge, Symbiotic-MoE noticeably enhances text-centric perception.

\begin{table}[t!]
\caption{\textbf{Additional Understanding Capabilities.} To ensure no implicit degradation occurs in specialized understanding domains, we expand our evaluation across diverse multimodal benchmarks, including spatial reasoning (POPE), compositional visual QA (GQA), chart and diagram understanding (ChartQA, AI2D), and comprehensive multimodal evaluations (TQA, MME). For the MME benchmark, we report the Perception score (MME-P), Cognition score (MME-C), and their total sum (MME-S). Symbiotic-MoE consistently preserves robust understanding across all domains compared to the Standard MoE and MoT baselines.}
\label{tab:supp_understanding}
\centering
\resizebox{0.90\textwidth}{!}{
\begin{tabular}{l|cccccccc}
\toprule
    Method & POPE$\uparrow$ & GQA$\uparrow$ & TQA$\uparrow$ & ChartQA$\uparrow$ & AI2D$\uparrow$ & MME-S$\uparrow$ & MME-P$\uparrow$ & MME-C$\uparrow$ \\
    \midrule
    \textcolor{gray}{Only\_LM\_MMU} & \textcolor{gray}{70.1} & \textcolor{gray}{45.2} & \textcolor{gray}{61.9} & \textcolor{gray}{66.8} & \textcolor{gray}{0.65} & \textcolor{gray}{1689.6} & \textcolor{gray}{1250.0} & \textcolor{gray}{439.6} \\
    \midrule
    Standard MoE & 60.3 & 38.5 & 55.9 & 57.5 & 0.56 & 1477.5 & 1093.8 & 383.7 \\
    MoT          & 63.1 & 40.9 & 57.5 & 60.1 & 0.57 & 1521.1 & 1125.4 & 395.7 \\
    \textbf{Symbiotic-MoE} & \textbf{74.5} & \textbf{48.0} & \textbf{66.9} & \textbf{70.3} & \textbf{0.67} & \textbf{1795.2} & \textbf{1328.1} & \textbf{467.1} \\
\bottomrule
\end{tabular}
}
\vspace{-0.5em}
\end{table}

\medskip\noindent\textit{General Perception, Reasoning, and Hallucination} (\eg, MME~\cite{fu2023mme}, GQA~\cite{hudson2019gqa}, POPE~\cite{li2023evaluating}): While naive co-training often exacerbates object hallucinations due to generative domain shifts, Symbiotic-MoE maintains robust structural alignment. It effectively resists the stability-plasticity dilemma, avoiding the fabrication of non-existent objects (validated by POPE) while excelling in compositional scene understanding (GQA) and comprehensive perception-reasoning (MME-P/MME-C).

\medskip\noindent\textit{Chart and Scientific Comprehension} (\eg, ChartQA~\cite{masry2022chartqa}, AI2D~\cite{kembhavi2016diagram}): These tasks demand rigorous numerical reasoning and topological understanding. The preservation of high scores in these domains demonstrates that our Modality-Aware Disentanglement successfully shields the foundational reasoning pathways from the high-variance gradients of the diffusion process. 

The collective results across these diverse taxonomies vividly illustrate the superiority of our framework. Standard MoE suffers from severe catastrophic forgetting due to unconstrained gradient interference. Conversely, while MoT merely preserves baseline capabilities via strict structural isolation, it fundamentally fails to leverage cross-modal synergy. In striking contrast, Symbiotic-MoE consistently mitigates forgetting and aligns closely with and notably surpasses the original VLM's only training on LM and MMU performance across all evaluated domains. The absence of a performance drop on these reasoning-heavy tasks, coupled with competitive generation scores, empirically corroborates our core claim: the symbiotic architecture successfully transforms generative signals into a constructive regularizer rather than a destructive interference.

\begin{figure}[t]
    \centering
    \begin{subfigure}{0.325\textwidth}
        \includegraphics[width=\linewidth]{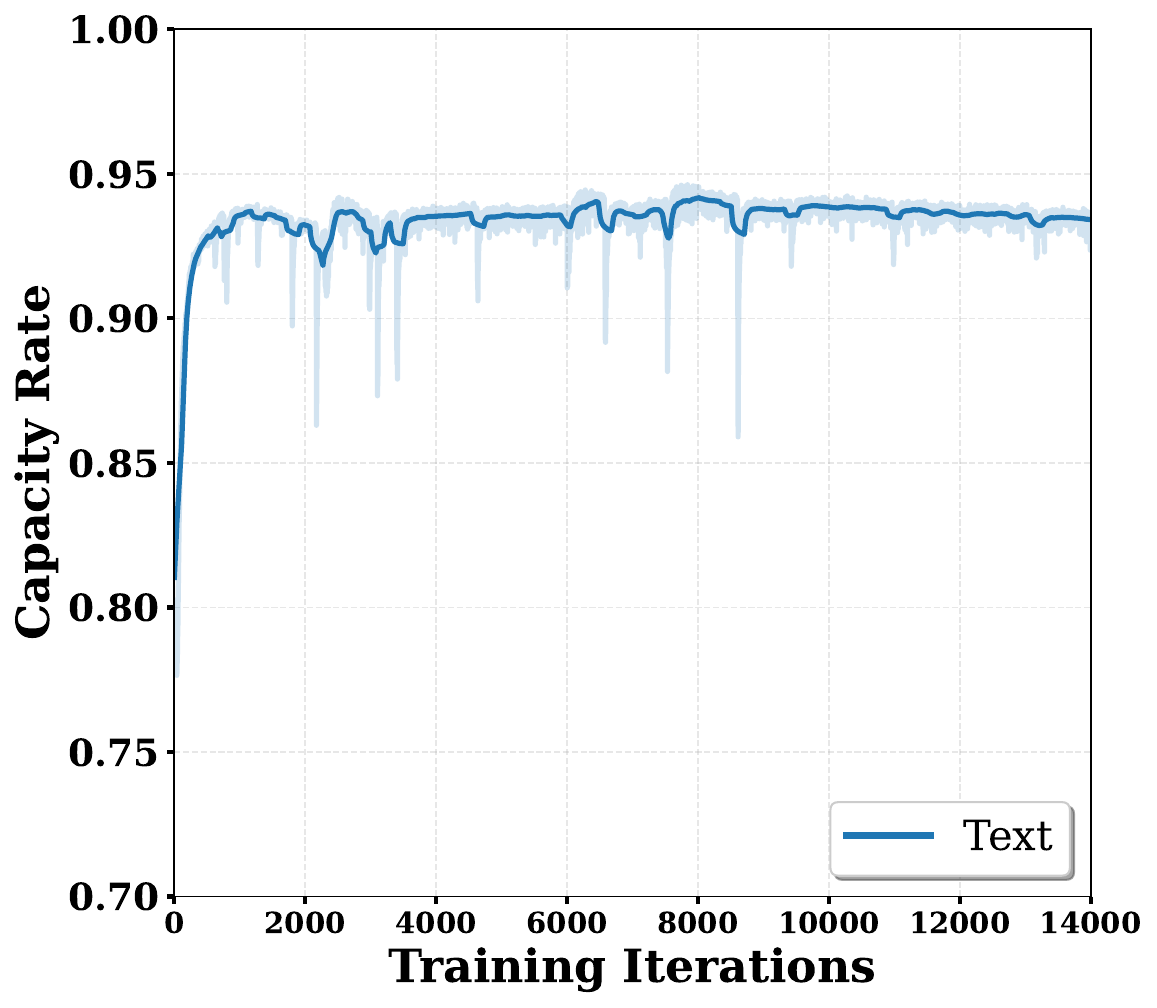} 
        \caption{Text}
        \label{fig:cap_text}
    \end{subfigure}
    \hfill
    \begin{subfigure}{0.325\textwidth}
        \includegraphics[width=\linewidth]{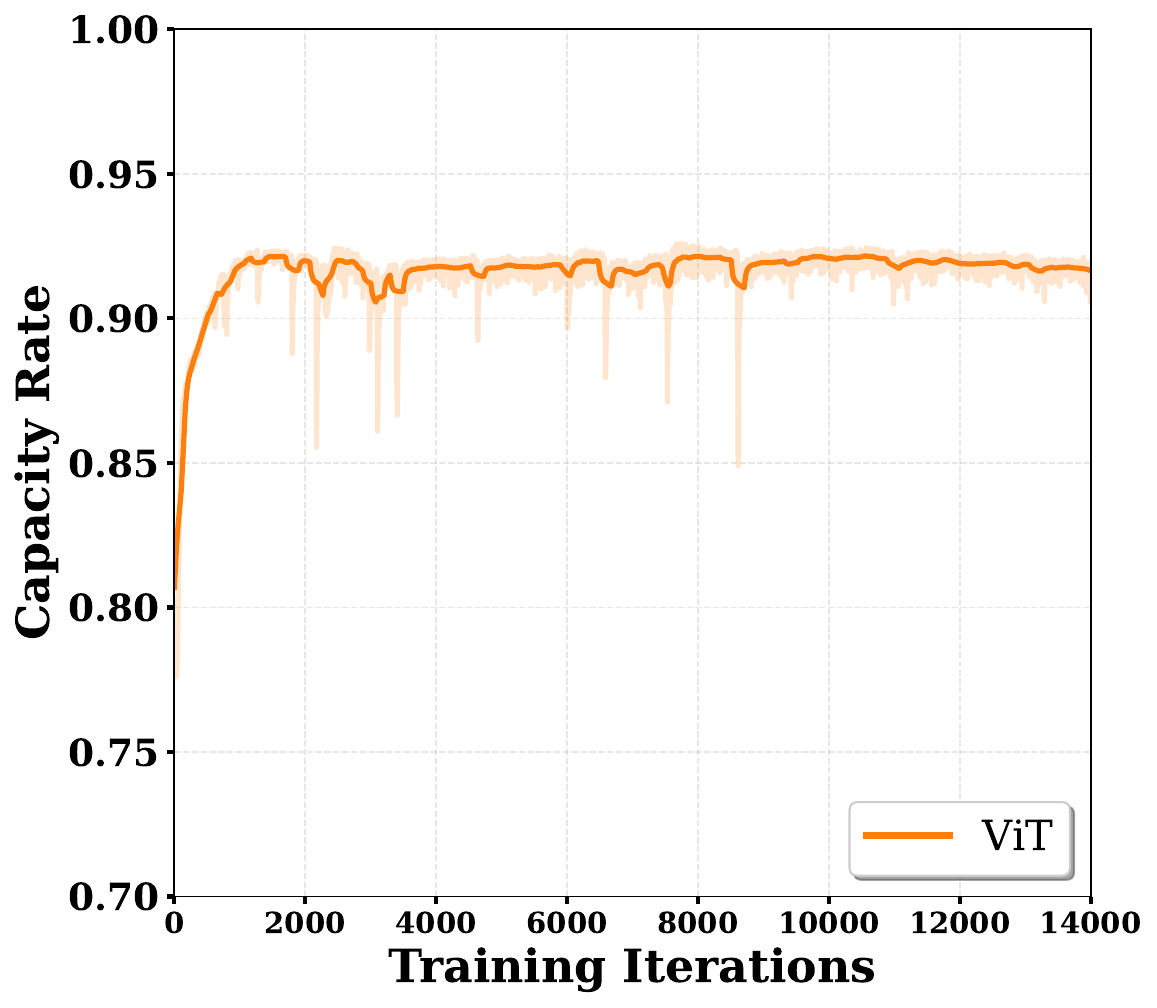}
        \caption{ViT}
        \label{fig:cap_vit}
    \end{subfigure}
    \hfill
    \begin{subfigure}{0.325\textwidth}
        \includegraphics[width=\linewidth]{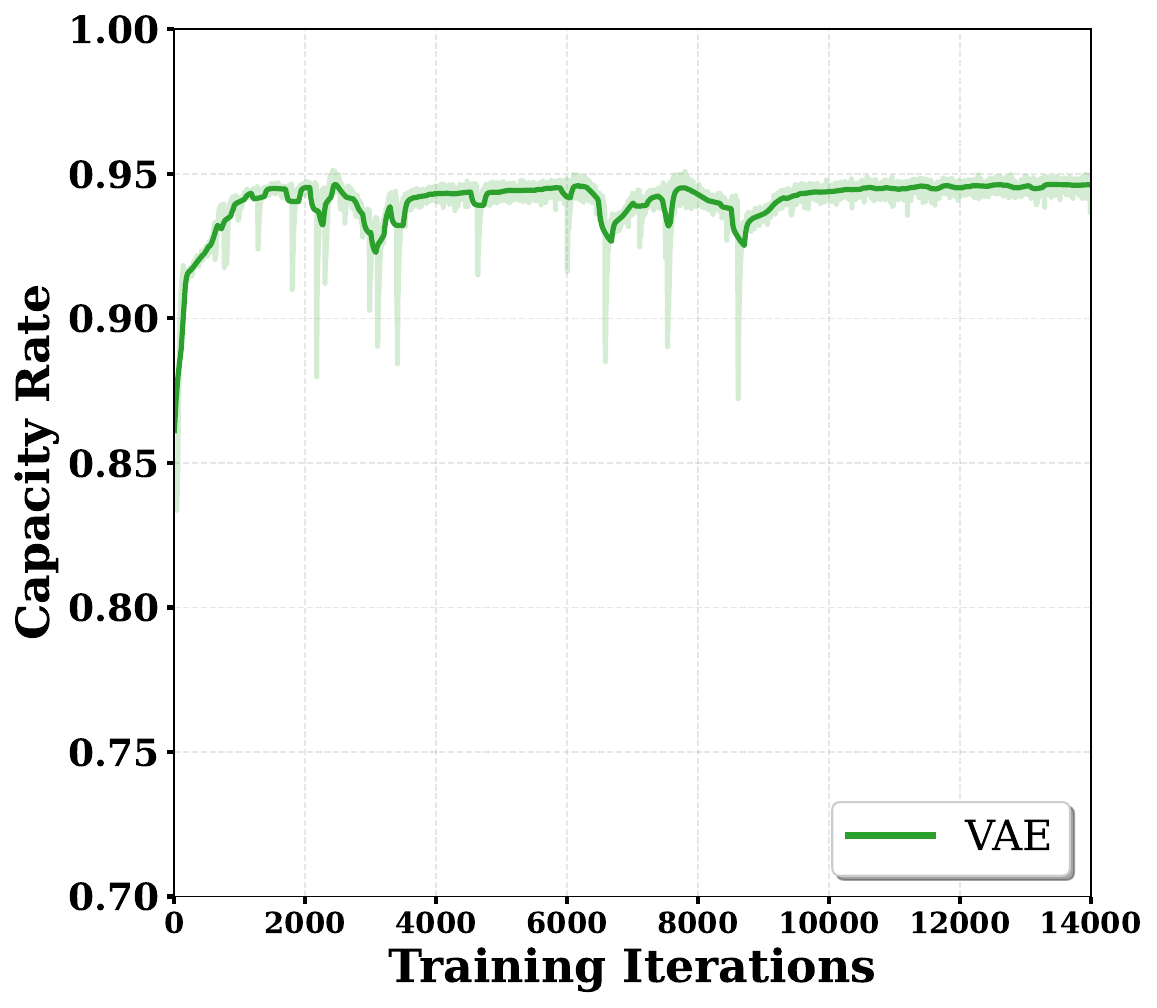}
        \caption{VAE}
        \label{fig:cap_vae}
    \end{subfigure}
    \caption{\textbf{Capacity Rate Dynamics of Text, ViT, and VAE.} We decompose the global capacity rate to verify modality-specific routing health. \textbf{(a-c)} All three modalities maintain an exceptionally high and stable capacity rate ($>0.90$), indicating near-perfect load balancing without partial routing collapse. The VAE curve closely mirrors the global trajectory (main paper Fig.~\ref{fig:capacity}) due to its dominant token volume in our $3:3:2:2$ data mixture, confirming the architectural robustness of Symbiotic-MoE across disparate modalities.}
    \label{fig:capacity_rate_each_modelity}
\end{figure}

\subsection{Fine-Grained Modality Routing Stability}
\label{subsec:fine_grained_dynamics}

In the main paper, we demonstrated that Symbiotic-MoE maintains a near-optimal global capacity rate ($\sim$0.95). To definitively rule out the possibility of partial routing collapse, where one dominant modality monopolizes expert utilization while others starve, we decompose the routing dynamics into modality-specific trajectories.

Figure~\ref{fig:capacity_rate_each_modelity} visualizes the individual capacity usage curves for Text, ViT, and VAE tokens throughout the training process. Strikingly, despite the disparate optimization paces enforced by our differential learning rates, all three modalities maintain exceptionally stable utilization above $0.90$, a highly ideal state in sparse MoE training, with Text and VAE consistently approaching $0.95$. 

This fine-grained visualization definitively rules out partial routing collapse and clarifies two key training dynamics. First, the global capacity curve presented in the main text visually mirrors the VAE trajectory simply because dense generative tokens constitute the absolute majority ($60\%$) of our training budget (\texttt{T2I: T2I-Long: LM: MMU = 3:3:2:2}), as corroborated by the token consumption in Fig.~\ref{fig:token_consumption}. Second, the marginally lower rate of the ViT branch ($\sim$0.93) is a natural consequence of the sparser routing signals from the MMU task, which provides the smallest token influx. Ultimately, this confirms that our near-optimal global efficiency ($\sim$0.95) is not a statistical artifact masked by the dominant modality, but a genuine reflection of system-wide load balancing successfully enforced by our Modality-Aware Disentanglement.

\section{Implementation Details}
\label{sec:supp_impl}
To ensure a rigorous comparison, our training protocol aligns strictly with the initial settings of the original VLM backbone. By isolating confounding variables, we guarantee that any performance variations are exclusively attributable to our architectural and optimization innovations.

\subsection{Training Paradigm and Evaluation Philosophy}
\label{sec:supp_impl_train}

We build upon the powerful \textbf{Hunyuan-VL-30B-A3B} backbone, adopting the Transfusion~\cite{zhou2024transfusion} framework to seamlessly unify the modeling of discrete text tokens and continuous visual latents. 

Modern foundation models typically undergo a multi-stage curriculum to achieve optimal aesthetic alignment: Pre-Training (PT), Continued Training (CT) for resolution scale-up, and Supervised Fine-Tuning (SFT) on curated high-quality subsets. In this work, our experiments are deliberately confined exclusively to the \textbf{PT Stage}.
Accordingly, the visual generation objective is restricted to a base pixel budget equivalent to a $256 \times 256$ resolution. Rather than forcing rigid square crops, we employ a dynamic aspect ratio bucketing strategy, where images are grouped into variable resolution buckets ($H \times W \approx 256^2$) to preserve their native semantic compositions. Note that we exclusively perform the Pre-Training (PT) phase from scratch, and no subsequent Continuous Training (CT) or Supervised Fine-Tuning (SFT) is employed in all of our experiments.

This constrained setting is a conscious experimental design choice. Our primary objective is to benchmark the \textit{intrinsic architectural superiority} of the Symbiotic-MoE in mitigating gradient conflicts and routing collapse, rather than blindly pursuing absolute aesthetic perfection via massive computational scaling. Because all baselines (Standard MoE and MoT) operate under the exact same resolution and data constraints, the relative performance margins and the robust preservation of understanding capabilities serve as definitive, unbiased evidence of our framework's foundational efficacy. Detailed hyperparameters for reproducibility are summarized in Table~\ref{tab:hyperparams}.

\begin{table}[t]
    \centering
    \caption{\textbf{Hyperparameter Settings.} The hyperparameters are strictly aligned with the pre-trained VLM to isolate the impact of our architectural improvements.
    }
    \label{tab:hyperparams}
    \small
    \scalebox{0.92}{
    \begin{tabular}{l|c}
        \toprule
        \textbf{Hyperparameter} & \textbf{Value} \\
        \midrule
        Base model & Hunyuan-VL-30B-A3B \\
        Learning rate (Gen) & $1 \times 10^{-4}$ \\
        Learning rate (Und) & $1 \times 10^{-6}$ \\
        LR scheduler & Constant \\
        Weight decay & 0.0 \\
        Gradient norm clip & 1.0 \\
        Optimizer & Adam ($\beta_1=0.9$, $\beta_2=0.95$, $\epsilon=10^{-6}$) \\
        Warm-up steps & 500 \\
        Gen resolution & $\approx 256 \times 256$ (bucketed) \\
        Loss Weight $\lambda_{disc}$ & 1.0 \\
        Loss Weight $\lambda_{img}$ & 1.0 \\
        Loss Weight $\lambda_{aux}$ & 0.01 \\
        Training iters & 30k \\
        Training seen tokens & 60B \\
        \bottomrule
    \end{tabular}
    }
\end{table}

\begin{figure}[t]
    \centering
    \includegraphics[width=0.45\textwidth]{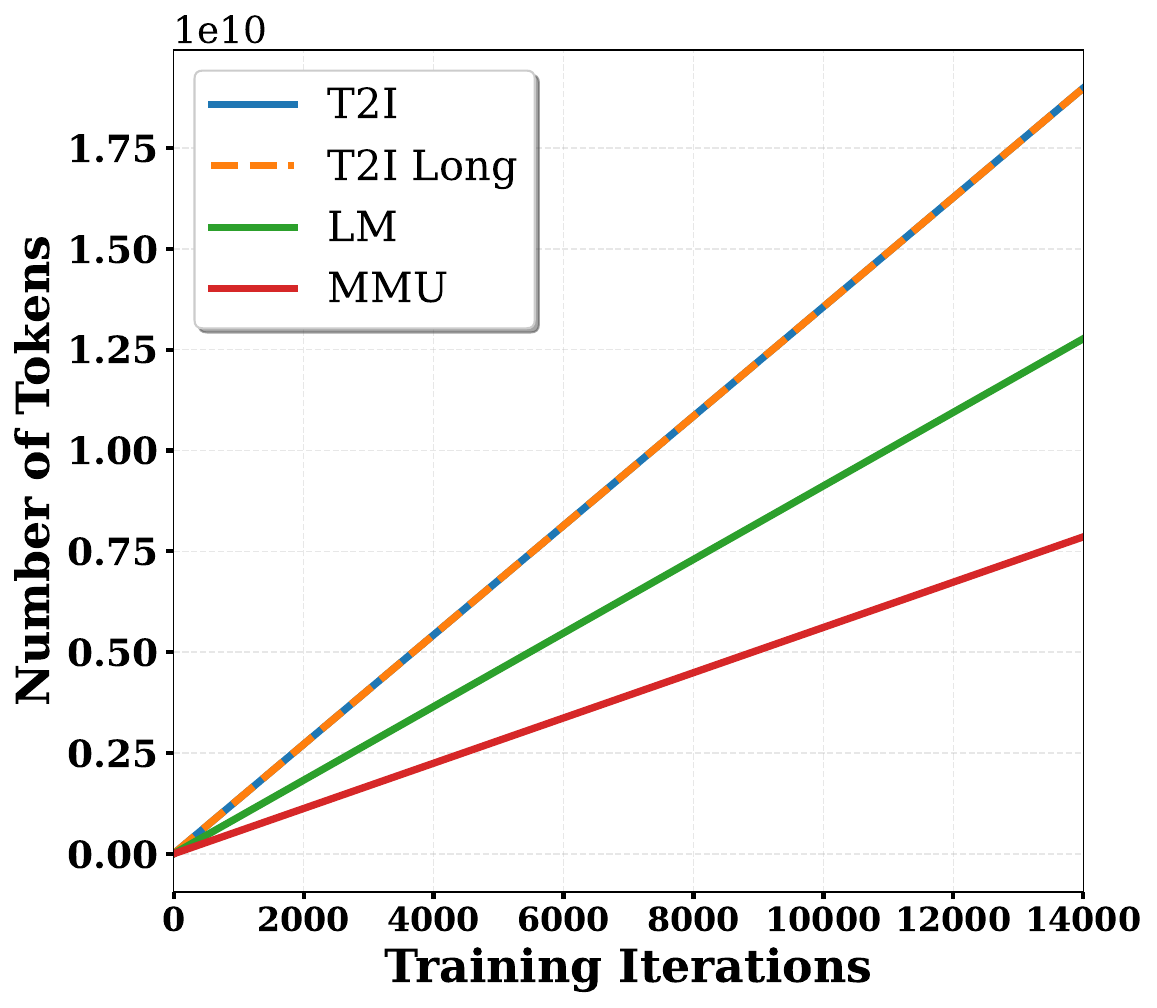}
    \caption{\textbf{Cumulative Token Consumption Dynamics.} We track the total number of tokens processed by the routing mechanism across 14,000 iterations. The massive influx of tokens across all three modalities highlights the rigorous scale of our co-training phase, providing a solid empirical foundation for cross-modal synergy.}
    \label{fig:token_consumption}
\end{figure}

\subsection{Dataset and Data Preparation}
\label{sec:data_details}

In this section, we provide the detailed dataset specifications and the data preparation pipeline used to construct our training corpus. In practice, our data curation, filtering, and captioning methods are inspired by HunyuanImage 3.0~\cite{cao2025hunyuanimage}.

\subsubsection{Data Scale and Three-Stage Filtering.} 

Following the data scale and filtering standards of HunyuanImage 3.0, we construct our pretraining dataset starting from a web-scale repository of over 10 billion raw image-text pairs. This massive pool is filtered down to approximately 4.5 billion high-quality samples (representing less than 45\% of the initial data). For our 100K-step training phase, we further subsample a representative subset of approximately 250 million high-quality pairs to optimize training efficiency and model convergence. The curation process follows a rigorous three-stage pipeline.

\paragraph{Technical and Quality Filtering.} In the first stage, we apply rule-based technical filters to discard low-resolution images ($<$ 512 pixels), corrupted files, and samples with extreme exposure or over-saturation, followed by MD5 deduplication. The second stage performs primary data curation using objective and subjective operators. For objective filtering, we deploy learning-based detectors to remove watermarks, corporate logos, photo collages, and prominent borders, alongside a high-accuracy OCR engine for text-dense layouts. To combat distribution shifts caused by synthetic datasets, we deploy an automated AI-Generated Content (AIGC) classifier combined with source-level domain blocking for web sources with a high ratio of generated images. For subjective curation, we evaluate image clarity (dynamic range, noise, and sharpness) and aesthetics using models co-designed with professional artists to establish unified filtering thresholds.

\paragraph{Deduplication and Sequential Mining.} The final stage implements semantic deduplication via embedding clustering, removing about 0.5\% of redundant data, and enriches the dataset with knowledge-augmented, stylized, and graphic design collections. To support interleaved learning, we also curate over 100 million image pairs and multi-image sequences. These are mined through image clustering (extracting highly similar pairs validated by a relation discriminator while removing overly complex element layouts) and video keyframe mining (using shot boundary detection, camera motion classification to exclude rapid transformations, object detection, and motion blur filtering).

\subsubsection{Hierarchical Captioning and Factual Grounding.}

To generate rich, controllable, and factually grounded descriptions, we propose an advanced image captioning pipeline. We inherit the core design of HunyuanImage 3.0, consisting of hierarchical schemas, compositional synthesis, and specialized grounding agents.

\paragraph{Hierarchical and Compositional Schema.} We establish a structured bilingual (English/Chinese) captioning schema that decomposes image semantics into distinct fields. This includes (i) \textit{descriptive levels} spanning four granularities from a concise summary to an exhaustive depiction of foreground/background elements, (ii) \textit{stylistic attributes} covering cinematography, lighting, and composition, and (iii) \textit{factual entities} identifying specific IPs, characters, landmarks, and brands. To enhance generalization and mitigate overfitting, we dynamically sample and synthesize different semantic fields of our hierarchical schema during training, producing augmented captions ranging from 30 to 1,000 words.

\paragraph{Factual Grounding and Verification.} Standard VLMs often hallucinate in-image text or fail at entity recognition. To enforce factual grounding, we integrate an OCR Agent to transcribe text and an IP/Entity Agent to identify real-world entities. This auxiliary information is fed to the captioning model as context. Crucially, we implement a Bidirectional Verification Loop that cross-references the detected entities with the generated caption; only samples successfully passing this bidirectional validation are retained in the final training set.

\subsection{Data Mixture and Evaluation Protocols}

To foster true cross-modal synergy, our training corpus comprises a high-quality, proprietary mixture of text, image-text pairs, and interleaved multimodal data. The sampling ratio during the co-training phase is deterministically set to \texttt{T2I : T2I-Long : LM : MMU = 3:3:2:2}. This balanced distribution ensures that the generative experts receive sufficient optimization signals without starving the discriminative understanding pathways.

To rigorously validate this balance, we adopt a comprehensive, dual-faceted evaluation protocol. For \textit{Generative Fidelity and Alignment}, we utilize GenEval~\cite{ghosh2023geneval}, FID~\cite{heusel2017gans}, CLIPScore~\cite{hessel2021clipscore}, and HPSv2~\cite{wu2023human} on COCO-30K~\cite{lin2014microsoft}, along with T2I-CompBench~\cite{huang2023t2i} for compositional semantics. Conversely, to verify \textit{Understanding Preservation}, we deploy a broad suite of reasoning and perception benchmarks, including MMLU~\cite{hendrycks2020measuring} (general knowledge), OCRBench~\cite{liu2024ocrbench} (fine-grained perception), POPE~\cite{li2023evaluating}, GQA~\cite{hudson2019gqa}, TQA~\cite{kembhavi2017you}, ChartQA~\cite{masry2022chartqa}, AI2D~\cite{kembhavi2016diagram}, and MME~\cite{fu2023mme}. This holistic approach ensures that the model's capabilities are evaluated across the entire spectrum of unified intelligence.

\subsection{Computational Budget and Token Dynamics}

All experiments are conducted on a high-performance cluster equipped with 256 NVIDIA H20 GPUs. Benefiting from the zero-parameter overhead of our Symbiotic-MoE architecture, the training process maintains optimal hardware utilization without introducing memory bottlenecks. The main model is trained for a total of 30,000 iterations using the AdamW optimizer. Given the massive Global Batch Size (GBS) of approximately 2,500 samples ($\sim$2M tokens) per iteration, the entire adaptation phase consumes roughly 60 Billion tokens.

\begin{figure}[!ht]
    \centering
    \includegraphics[width=1.0\textwidth]{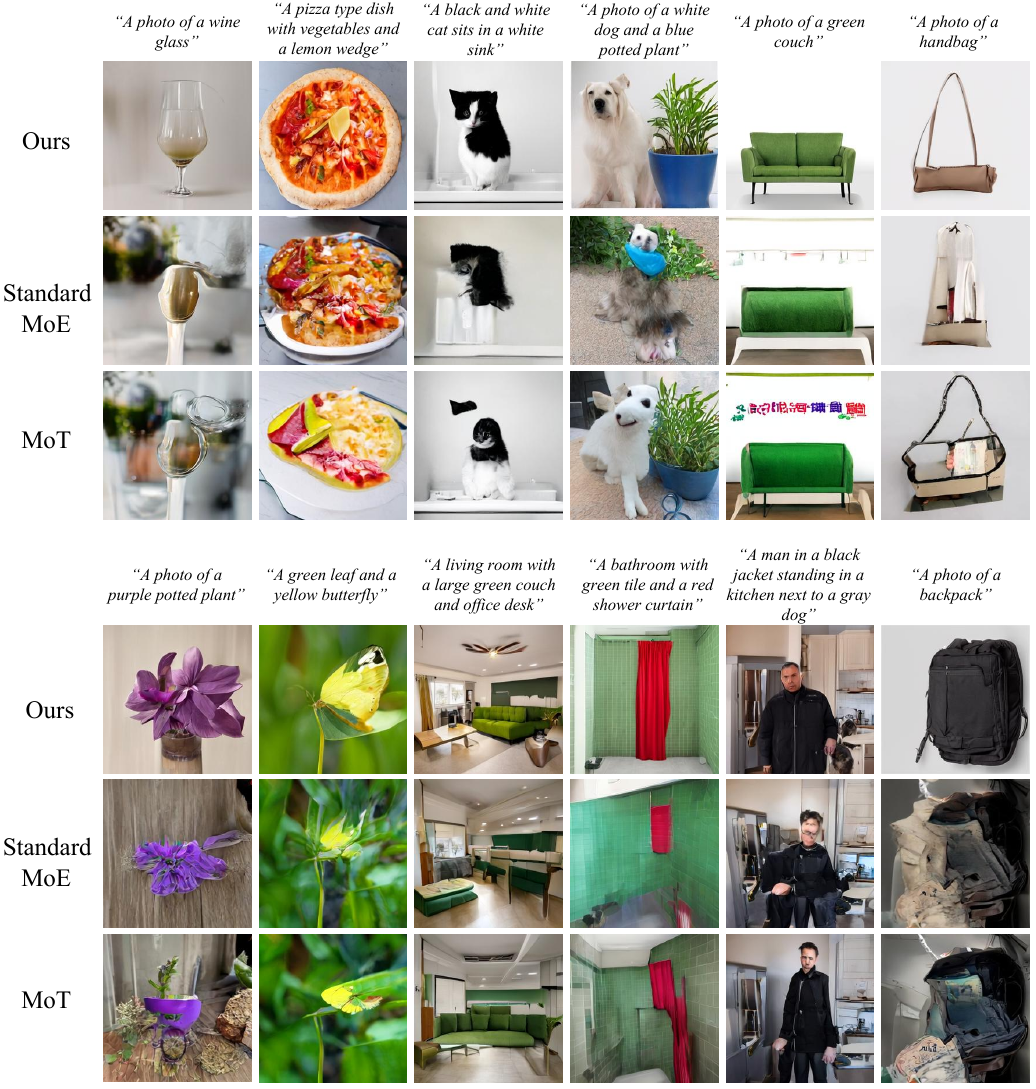}
    \caption{\textbf{Extended Qualitative Comparison.} 
    % We compare the text-to-image generation capabilities of Symbiotic-MoE against the Standard MoE and MoT baselines. Standard MoE suffers from catastrophic gradient interference, leading to complete structural collapse (\eg, the handbag and backpack). MoT, constrained by its physically isolated ``split-brain'' architecture, lacks deep semantic grounding, resulting in severe attribute misalignment and structural hallucinations (\eg, the missing couch legs, anomalous artifacts, and distorted wine glass). In contrast, Symbiotic-MoE leverages shared experts as a semantic bridge, consistently synthesizing high-fidelity images with precise compositional attribute binding and coherent geometry.
    We compare the early-stage text-to-image generation capabilities of Symbiotic-MoE against the Standard MoE and MoT baselines. At this stage of training, Standard MoE struggles to form coherent structures due to gradient interference (\eg, the handbag). While MoT isolates these conflicts, it exhibits slower semantic alignment, leading to occasional attribute mismatches (\eg, the wine glass). In contrast, Symbiotic-MoE leverages shared experts to accelerate cross-modal convergence, consistently synthesizing higher-fidelity images with more precise compositional attribute binding earlier in the training process.
    }
    \label{fig:supp_comparison}
\end{figure}

To provide a granular perspective on this computational scale, Fig.~\ref{fig:token_consumption} illustrates the cumulative token consumption throughout the training trajectory. Consistent with our \texttt{T2I: T2I-Long: LM: MMU = 3:3:2:2} data mixture and the inherently high sequence density of continuous visual latents, VAE tokens dominate the overall volume. Concurrently, Text and ViT tokens scale steadily, providing a continuous regularization effect for the understanding modules. This massive and modality-proportional influx of data guarantees that both the newly initialized generative experts and the pre-trained understanding anchors reach a state of robust convergence.

% =======================================================
% Section D: Visuals
% =======================================================
\section{Additional Qualitative Results}
\label{sec:supp_vis}

In this section, we provide extended qualitative visualizations to further validate the generative superiority of Symbiotic-MoE. Figure~\ref{fig:supp_comparison} presents a side-by-side comparison against the Standard MoE and MoT baselines, highlighting our model's capacity to avert structural collapse and maintain precise semantic alignment. 

\medskip\noindent\textbf{Structural Integrity and Fine-grained Details.} 
Standard MoE suffers from severe gradient interference, which irreversibly corrupts foundational visual priors. As observed in the ``handbag'' and ``wine glass'' prompts, Standard MoE completely collapses into amorphous textures. While MoT mitigates this collapse via physical isolation, it frequently hallucinates erroneous geometric structures, such as generating multiple floating rims for the wine glass or distorted limbs for the cat. In stark contrast, Symbiotic-MoE synthesizes geometrically precise objects with photorealistic details (e.g., the accurate refraction of the glass and the clean silhouette of the cat). This confirms that our optimization dynamics successfully protect generative plasticity from being overwhelmed by understanding tasks.

\medskip\noindent\textbf{Compositional Attribute Binding.} 
The most pronounced advantage of our Symbiotic-MoE emerges in prompts requiring complex attribute binding. Because MoT structurally isolates the text understanding experts from the generation pathways (the ``split-brain'' dilemma), its generative module lacks robust semantic grounding. Consequently, MoT struggles with precise compositional generation: it fails to render the ``lemon wedge'' on the pizza correctly, morphs the ``purple potted plant'' into an anomalous purple cart, and hallucinates random textual artifacts above the ``green couch'' while failing to generate its legs. 
Conversely, Symbiotic-MoE flawlessly renders the distinct lemon wedge, the delicate wings of the yellow butterfly, and the four distinct legs of the green couch. This precise alignment is directly attributable to our \textit{Shared Experts}, which act as a semantic bridge. By allowing the generative decoder to query deeply aligned textual representations without destroying them, Symbiotic-MoE ensures that fine-grained textual attributes are strictly bound to their corresponding visual entities.

\bibliographystyle{splncs04}
\bibliography{main}
\end{document}